%% file: main.tex
\documentclass[runningheads]{llncs}
\pdfoutput=1 

\usepackage{graphicx}
\usepackage{comment}
\usepackage{amsmath,amssymb}
\usepackage{color}
\usepackage{url}
\usepackage{hyperref}
\usepackage{subcaption}
\usepackage[super]{nth}
\usepackage{siunitx}
\usepackage{booktabs}
\usepackage{etoolbox}
\usepackage{enumitem}
\usepackage{multirow}
\usepackage{verbatim}
\usepackage{xcolor}
\usepackage{array}
\usepackage{xspace}

\usepackage[capbesideposition=outside,capbesidesep=quad]{floatrow}

\usepackage[toc,page]{appendix}

\newcommand{\boldparagraph}[1]{\vskip1pt \noindent{\bf #1~}}

\renewrobustcmd{\bfseries}{\fontseries{b}\selectfont}
\renewrobustcmd{\boldmath}{}
\newrobustcmd{\B}{\bfseries}

\newcolumntype{H}{>{\setbox0=\hbox\bgroup}c<{\egroup}@{}}

\newcommand{\gb}[1]{{\tiny\textcolor{darkgray}{#1}}}

\makeatletter
\DeclareRobustCommand\onedot{\futurelet\@let@token\@onedot}
\def\@onedot{\ifx\@let@token.\else.\null\fi\xspace}

\def\eg{\emph{e.g}\onedot} 

\def\ie{\emph{i.e}\onedot}

\def\etal{\emph{et al}\onedot}

\usepackage[capitalize]{cleveref}
\crefname{section}{Sec.}{Secs.}
\Crefname{section}{Section}{Sections}
\Crefname{table}{Table}{Tables}
\crefname{table}{Tab.}{Tabs.}

\def\shortTitle{NeuralMeshing}
\def\nir{NIR}
\def\nirs{NIRs}

\newif\ifreview
\reviewfalse

\begin{document}

\def\SubNumber{37}

\def\GCPRTrack{Main Track}

\title{\shortTitle{}: Differentiable Meshing\\ of Implicit Neural Representations}

\author{Mathias Vetsch\inst{1} \and
Sandro Lombardi\inst{1} \and
Marc Pollefeys\inst{1,2} \and\\
Martin R. Oswald\inst{1,3}}

\authorrunning{M. Vetsch et al.}

\institute{Department of Computer Science, ETH Zurich, Switzerland
\and Mixed Reality and AI Zurich Lab, Microsoft, Switzerland
\and University of Amsterdam, Netherlands
}

\maketitle
\input{symbols}

\input{00_abstract}

\input{01_introduction}
\input{02_related_work}
\input{03_method}

\input{04_evaluation}

\input{05_conclusion}

\boldparagraph{Acknowledgments.}
This work has been supported by Innosuisse funding (Grant No. 100.567 IP-ICT).

\clearpage
\bibliographystyle{splncs04}
\bibliography{egbib}

\clearpage

\centerline{\large\bf Appendix}%
\begin{appendix}
\input{supplementary_content}
\end{appendix}

\end{document}

%% file: symbols.tex
\newcommand{\abs}[1]{\left|#1\right|}
\newcommand{\norm}[1]{\left\lVert#1\right\rVert}
\newcommand{\clamp}{\ensuremath{\text{clamp}}} %
\newcommand{\R}{\mathbb{R}}

\newcommand{\XPC}{\ensuremath{\mathbf{P}}} %
\newcommand{\Xp}{\ensuremath{\mathbf{p}}} %
\newcommand{\Xn}{\ensuremath{\mathbf{n}}} %
\newcommand{\Xnumpoints}{\ensuremath{N}} %
\newcommand{\XNIP}{\ensuremath{\mathcal{S}}} %
\newcommand{\XM}{\ensuremath{\mathcal{M}}} %
\newcommand{\XV}{\ensuremath{\mathbf{V}}} %
\newcommand{\Xv}{\ensuremath{\mathbf{v}}} %
\newcommand{\XF}{\ensuremath{\mathbf{F}}} %
\newcommand{\Xf}{\ensuremath{\mathbf{f}}} %

\newcommand{\Xq}{\Xp} %
\newcommand{\Xangle}{\ensuremath{\phi}} %
\newcommand{\Xldist}{\ensuremath{l}} %
\newcommand{\Xqdir}{\ensuremath{\mathbf{q}}} %
\newcommand{\Xcurv}{\ensuremath{\kappa}} %

\newcommand{\Xnetw}{\ensuremath{\theta}} %
\newcommand{\Xlat}{\ensuremath{\mathbf{z}}} %
\newcommand{\Xnet}{\ensuremath{f}} %
\newcommand{\Xsdf}{\ensuremath{s}} %
\newcommand{\Xloss}{\ensuremath{\mathcal{L}}} %
\newcommand{\Xlossw}{\ensuremath{\lambda}} %

\newcommand{\Xrad}{\ensuremath{r}} %
\newcommand{\Xrd}{\ensuremath{\Xrad_{d}}} %
\newcommand{\Xtri}{\ensuremath{T}} %
\newcommand{\Xnumtriangles}{\ensuremath{I}} %
\newcommand{\Xdist}{\ensuremath{d}} %
\newcommand{\Xnumprojections}{\ensuremath{P}} %

\newcommand{\Xpred}{\ensuremath{\mathbf{r}}} %

\newcommand{\Xpreder}{\ensuremath{{r}_{\text{ER}}}} %
\newcommand{\Xpredls}{\ensuremath{{r}_{\text{LS}}}} %
\newcommand{\Xpredfr}{\ensuremath{{r}_{\text{FR}}}} %

\newcommand{\Xdvec}{\ensuremath{\mathbf{v}_{\text{d}}}} %
\newcommand{\Xpredvert}{\ensuremath{\Xp_{\text{d}}}} %
\newcommand{\Xbecent}{\ensuremath{\mathbf{m}}} %
\newcommand{\Xbestart}{\ensuremath{\mathbf{a}}} %
\newcommand{\Xbeend}{\ensuremath{\mathbf{b}}} %
\newcommand{\Xfnormal}{\ensuremath{\Xn_{\text{F}}}} %
\newcommand{\XbeSDFc}{\ensuremath{\Xsdf_{\Xpredvert}}} %
\newcommand{\XbeSDFs}{\ensuremath{\Xsdf_{\Xbestart}}} %
\newcommand{\XbeSDFe}{\ensuremath{\Xsdf_{\Xbeend}}} %
\newcommand{\XbeGradc}{\ensuremath{\Xn_{\Xpredvert}}} %
\newcommand{\XbeGrads}{\ensuremath{\Xn_{\Xbestart}}} %
\newcommand{\XbeGrade}{\ensuremath{\Xn_{\Xbeend}}} %
\newcommand{\XbeCurvc}{\ensuremath{\Xcurv_{\Xpredvert}}} %
\newcommand{\XbeCurvs}{\ensuremath{\Xcurv_{\Xbestart}}} %
\newcommand{\XbeCurve}{\ensuremath{\Xcurv_{\Xbeend}}} %

\newcommand{\Xlossigr}{\ensuremath{\Xloss_{\text{IGR}}}} %
\newcommand{\Xlosscurv}{\ensuremath{\Xloss_{\text{Curv}}}} %
\newcommand{\Xlosswcurv}{\ensuremath{\Xlossw_{\text{Curv}}}} %

\newcommand{\Xlosstot}{\ensuremath{\Xloss_{\text{Total}}}} %
\newcommand{\Xlossbe}{\ensuremath{\Xloss_{\text{ER}}}} %
\newcommand{\Xlosssd}{\ensuremath{\Xloss_{\text{SD}}}} %
\newcommand{\Xlossreg}{\ensuremath{\Xloss_{\text{LR}}}} %
\newcommand{\Xlosswbe}{\ensuremath{\Xlossw_{\text{ER}}}} %
\newcommand{\Xlosswsd}{\ensuremath{\Xlossw_{\text{SD}}}} %
\newcommand{\Xlosswreg}{\ensuremath{\Xlossw_{\text{LR}}}} %

\newcommand{\Xgtcurv}{\ensuremath{\kappa_{\text{GT}}}} %
\newcommand{\Xgtangle}{\ensuremath{\Xangle_{\text{GT}}}} %
\newcommand{\Xpredsdf}{\ensuremath{\Xsdf_{\Xpredvert}}} %

\newcommand{\eucspace}{\R^3}
\newcommand{\pointcloud}{\XPC}
\newcommand{\normals}{\mathcal{N}}
\newcommand{\latentcode}{\Xlat}
\newcommand{\decoderweights}{\Xnetw}
\newcommand{\defaultprediction}{\Xdvec}
\newcommand{\unitboundaryedge}{\mathbf{e}_u}
\newcommand{\facenormal}{\Xfnormal}

\newcommand{\scalaredgerotation}{\Xpreder}
\newcommand{\scalarfacenormalotation}{\Xpredfr}
\newcommand{\scalarlengthscaling}{\Xpredls}

\newcommand{\defaultradius}{\Xrd}
\newcommand{\pihalbe}{\frac{\pi}{2}}

%% file: 00_abstract.tex
\begin{abstract}
The generation of triangle meshes from point clouds, \ie meshing, is a core task in computer graphics and computer vision.
Traditional techniques directly construct a surface mesh using local decision heuristics, while some recent methods based on neural implicit representations try to leverage data-driven approaches for this meshing process.
However, it is challenging to define a learnable representation for triangle meshes of unknown topology and size and for this reason, neural implicit representations rely on non-differentiable post-processing in order to extract the final triangle mesh.
In this work, we propose a novel differentiable meshing algorithm for extracting surface meshes from neural implicit representations.
Our method produces the mesh in an iterative fashion, which makes it applicable to shapes of various scales and adaptive to the local curvature of the shape.
Furthermore, our method produces meshes with regular tessellation patterns and fewer triangle faces compared to existing methods.
Experiments demonstrate the comparable reconstruction performance and favorable mesh properties over baselines.
\keywords{Meshing  \and Deep learning.}
\end{abstract}

%% file: 01_introduction.tex
\section{Introduction}
\label{sec:intro}
Meshing of 3D point clouds has been studied extensively.
Traditional methods either employ direct approaches based on 
local neighborhood properties
\cite{edelsbrunner_shape_1983,edelsbrunner_threedimensional_1994,boissonnat_geometric_1984,amenta_new_1998,amenta_power_2001,amenta_power_2001a,amenta_surface_1999} or use an implicit volumetric representation as an intermediary step~\cite{hoppe_surface_1992,carr_reconstruction_2001,lewiner_efficient_2003,lorensen_marching_1987,ju_dual_2002,schaefer_dual_2004,taubin_smooth_2012,fleishman_robust_2005,kolluri_provably_2008,oztireli_feature_2009,kazhdan_poisson_2006,kazhdan_screened_2013}. 
While early works perform poorly on noisy real-world input or exhibit high computational demands, follow-up methods have addressed several of these shortcoming~\cite{lv_voxel_2022,thayyil_local_2021,zhong_surface_2019,boltcheva_surface_2017}.

\begin{figure}[t]
    \centering
    \includegraphics[width=1.0\textwidth]{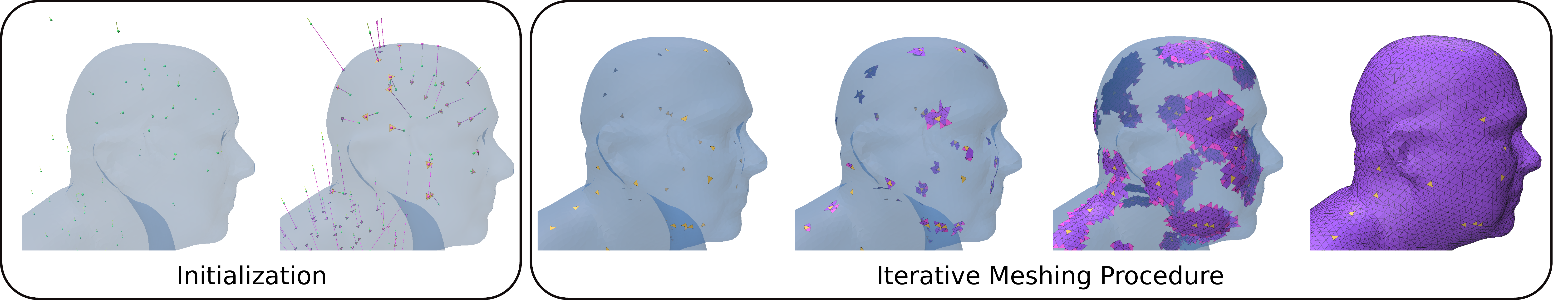}\\[-6pt]
    \caption{\textbf{\shortTitle{}.} We propose a novel meshing algorithm specifically designed for neural implicit representations (NIRs). Starting from an initial set of randomly placed seed triangles on the zero level set of the implicit neural representation, \shortTitle{} iteratively expands the triangles into all directions until the full zero level set is covered}
    \label{fig:teaser}
\end{figure}

In recent years, neural implicit representations (NIRs)~\cite{park_deepsdf_2019,mescheder_occupancy_2019,michalkiewicz_deep_2019,chen_learning_2018} have been used to improve upon traditional implicit-based representations by storing the implicit function within a deep neural network.
Follow-up approaches based on these have addressed various issues, \eg, concerning scalability~\cite{peng_convolutional_2020,chabra_deep_2020,jiang_local_2020}, quality~\cite{duan_curriculum_2020,sitzmann_implicit_2020,sitzmann_metasdf_2020} or processing of raw data~\cite{atzmon_sal_2020,atzmon_sald_2020,gropp_implicit_2020}.
However, such methods still rely on an isosurface extraction method like marching cubes~\cite{lewiner_efficient_2003,lorensen_marching_1987} in order to generate the final triangle mesh, usually resulting in unnecessarily high-resolution triangle meshes.
Furthermore, those post-processing steps are often not differentiable, prohibiting end-to-end training of networks.

In this paper, we propose \shortTitle{} (Figure~\ref{fig:teaser}), a novel data-driven approach for directly predicting a triangle-mesh from a \nir{}.
\shortTitle{} aims to close the gap of a differentiable meshing approach specifically designed for the usage with \nirs{}.
Starting from a seed of initially placed triangles, \shortTitle{} iteratively extends triangles at boundary edges by predicting new vertex locations given local geometry information like curvature, SDF and surface normals through queries on the underlying implicit representation.
This allows to adaptively place bigger triangles at surface areas with lower curvature, \ie flat surface patches, and smaller triangles at areas with high curvature.
The main contributions of this paper can be summarized as follows:
\textbf{(1)} We propose \shortTitle{}, a novel data-driven meshing approach for \nirs{}. Our method iteratively predicts new triangles based on local surface information from the implicit representation.
\textbf{(2)} Extensive experiments show that \shortTitle{} better approximates the surface of \nirs{} while using considerably fewer triangles than commonly used iso-surface extraction methods~\cite{lewiner_efficient_2003}.

%% file: 02_related_work.tex
\section{Related Work}
\label{sec:related_work}
\boldparagraph{Traditional Deterministic Reconstruction Methods.}
Early works on shape reconstruction from points have proposed deterministic approaches using alpha shapes~\cite{edelsbrunner_shape_1983,edelsbrunner_threedimensional_1994}, Delaunay triangulation~\cite{boissonnat_geometric_1984} or ball-pivoting~\cite{bernardini_ballpivoting_1999}.
Such methods make local decisions to directly triangulate the given input point cloud.
Later works focused on extracting the surface as the crust of a Voronoi diagram~\cite{amenta_new_1998,amenta_power_2001,amenta_power_2001a,amenta_surface_1999}.
However, these methods do not well handle noise or outliers and thus perform poorly on real-world data, creating noisy and incomplete output meshes.

\boldparagraph{Traditional Implicit Reconstruction Approaches.}
In contrast to triangulation methods, implicit-based approaches try to represent the surface as an implicit function.
The pioneering work of Hoppe~\etal~\cite{hoppe_surface_1992} introduced a method for creating a piecewise smooth surface through implicit modeling of a distance field.
An alternative approach relies on radial basis function methods~\cite{carr_reconstruction_2001} which try to fit implicit functions to a surface.
Another line of work focuses on extracting the iso-surface from signed distance function values (SDF) of a volumetric grid, the most prominent known as the marching cubes algorithm~\cite{lorensen_marching_1987,lewiner_efficient_2003}. 
There has been a plethora of follow-up work~\cite{ju_dual_2002,taubin_smooth_2012,schaefer_dual_2004}, improving upon marching cubes.
Moving least-squares (MLS)~\cite{fleishman_robust_2005,kolluri_provably_2008,oztireli_feature_2009} based techniques locally reconstruct the surface with local functions approximating the SDF in the local neighborhood.
Poisson surface reconstruction~\cite{kazhdan_poisson_2006,kazhdan_screened_2013} reconstructs surfaces from oriented points via energy optimization. 
While these methods are able to close larger surface holes they come with high computational demands.

\boldparagraph{Neural Implicit Representations.}
Recently, neural implicit representations are used as an alternative representation for surface reconstruction. Pioneering works \cite{chen_learning_2018,mescheder_occupancy_2019,michalkiewicz_deep_2019,park_deepsdf_2019} use coordinate-based deep networks in order to learn a continuous SDF or occupancy function. While the early works have been limited to objects and low levels of details, follow-up approaches have extended the representations to scenes of larger scale~\cite{chabra_deep_2020,chibane_implicit_2020,jiang_local_2020,lombardi_scalable_2020,mi_ssrnet_2020,peng_convolutional_2020}, proposed learning on raw data~\cite{atzmon_sal_2020,atzmon_sald_2020,chibane_neural_2020,gropp_implicit_2020}, improved details~\cite{sitzmann_implicit_2020} or improved upon the training scheme~\cite{duan_curriculum_2020,sitzmann_metasdf_2020}.
In order to obtain the final mesh, all these methods rely on extracting the surface via an iso-surface extraction approach like marching cubes~\cite{lorensen_marching_1987}.

\boldparagraph{Data-driven Direct Meshing Approaches.}
Recent works have started to adopt deep learning-based approaches for triangulation and meshing of shapes.
Early approaches used deep networks in order to warp 2D patches onto the point cloud~\cite{groueix_atlasnet_2018,williams_deep_2019}. Such parametric approaches often lead to undesirable holes and overlapping patches.
Some works proposed to utilize grids, \eg to predict a signed distance field using random forests~\cite{ladicky_point_2017}, to deform and fit a shape~\cite{yang_foldingnet_2018} or to extract the surface via a differentiable marching cubes algorithms~\cite{liao_deep_2018}. However, they come with high computational resource demands.
BSP-Net~\cite{chen_bspnet_2020} and CvxNet~\cite{deng_cvxnet_2020} both build upon the idea of predicting a set of hyperplanes for creating convex shapes as building blocks of the final shape. The triangulation is extracted through a non-differentiable post-processing step involving convex hull computations in the dual domain.
However, the number of planes is fixed which limits the reconstruction to objects of smaller scale.
Another line of work deforms and fits a template mesh to the input point cloud~\cite{hanocka_point2mesh_2020,pan_deep_2019,wang_pixel2mesh_2018,wen_pixel2mesh_2019}.
However, the topology of the reconstructed shape is usually fixed to the topology of the provided template mesh.
Recently, PointTriNet~\cite{sharp_pointtrinet_2020} proposed to directly predict the connectivity of the given input point cloud. 
In an iterative procedure, triangle candidates are first proposed and then classified for their suitability to be part of the final mesh. 
While the method is local and differentiable, the resulting meshes often include holes. Similarly, Rakotosaona~\etal~\cite{rakotosaona_learning_2021} model the problem of triangulation locally by predicting a logarithmic map which allows triangulation in 2D using non-differentiable Delaunay triangulation. 
Finally, some recent work focus on using the neural implicit representation as the core representation for differentiable meshing or learning~\cite{chen_neural_2021,peng_shape_2021,guillard_deepmesh_2021}.
Neural Marching Cubes~\cite{chen_neural_2021} implements a data-driven approach to improve the mesh quality at sharp edges by defining a learnable representation to differentiate between the topological configurations of marching cubes.
They introduce additional vertices inside cells of the underlying grid and predict the offset of these vertices.
However, they use a 3D ResNet and rely on discretized inputs, limiting the resolution and making it less memory efficient. 
In contrast, our method operates in continuous space and therefore on arbitrary resolutions.
DeepMesh~\cite{guillard_deepmesh_2021} uses a trick, \ie, an additional forward pass on all mesh vertex locations for computing gradients with respect to an underlying implicit representation without the need to make the meshing differentiable. 
They use a non-differentiable marching cubes algorithm to generate the output and define loss functions directly on the obtained mesh.
The %
recent work by Peng~\etal~\cite{peng_shape_2021} instead proposes a new shape representation based on a differentiable Poisson solver.
Contrary to those two works, we aim to directly create a triangle mesh from the underlying neural implicit representation.

%% file: 03_method.tex
\section{Method}
\label{sec:method}

\begin{figure}[tb]
    \centering
    \includegraphics[width=1.0\linewidth]{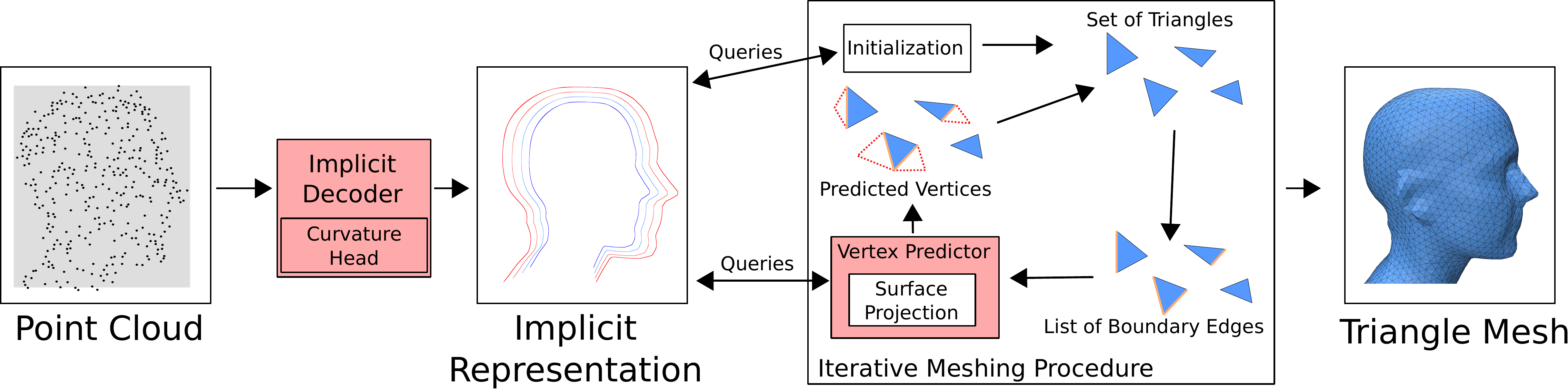}\\[-5pt]
    \caption{\textbf{Overview.} Given an input point cloud $\XPC$ (left), we use a neural implicit representation with added curvature head $\XNIP$ (middle) to extract a continuous SDF field. After placing a set of random triangles on the zero level set of $\XNIP$, \shortTitle{} then queries $\XNIP$ in order to predict a triangular mesh $\XM$ (right) in an iterative fashion. Red blocks denote trainable MLPs
    }
    \label{fig:overview}
\end{figure}

\shortTitle{} takes as input an oriented point cloud $\XPC = \{(\Xp_{i}, \Xn_{i})\ |\ \Xp_{i}\in\R^3, \Xn_{i}\in\R^3\}_{i=1}^{\Xnumpoints}$.
Our goal is to compute a surface mesh $\XM = (\XV, \XF)$ defined as a set of vertices $\XV =\{\Xv_i \in \R^3\}$ and a set of triangular faces $\XF$ %
in order to approximate the surface of a shape.
In a first step, we employ a modified neural implicit representation $\XNIP$ in order to learn a continuous SDF field approximating the shape's surface.
In a second step, we use 
the neural representation %
as input for our meshing algorithm in order to extract an explicit representation $\XM$ of the surface.
In order to effectively predict new triangles, our neural representation $\XNIP$ uses an extra branch in addition to the existing SDF branch, which outputs curvature information. 
Please refer to Figure~\ref{fig:overview} for an overview of our method.
\subsection{Modified Neural Implicit Representation}
Ideally, we aim for small triangles where the curvature is high and larger triangles at low curvature.
To account for this, we extend the neural implicit representation $\XNIP$ with curvature information.
In order to learn a \nir{}, we follow Gropp~\etal~\cite{gropp_implicit_2020} and use implicit geometric regularization (IGR) for network training.

\begin{figure}[tb]
    \captionsetup[subfigure]{aboveskip=0pt,belowskip=-5pt}
	\centering
	\begin{subfigure}[t]{0.25\linewidth}
		\centering
        \includegraphics[width=1.0\linewidth]{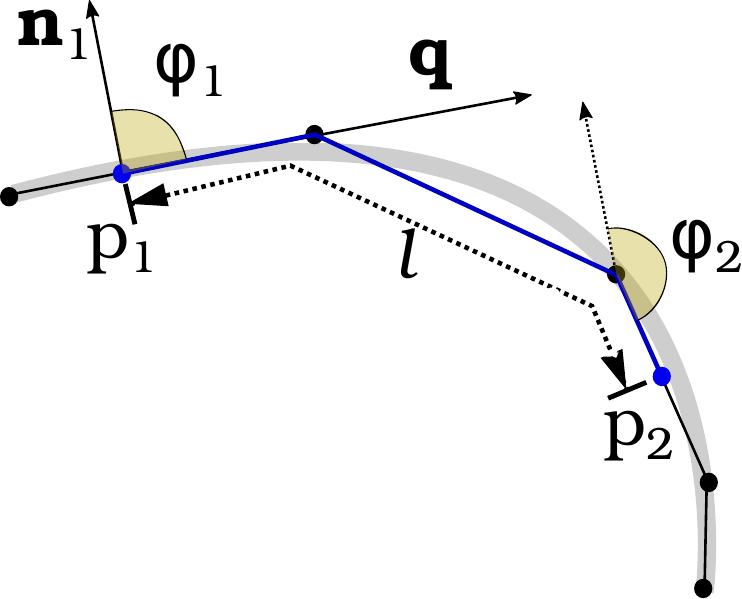}
        \caption{\textbf{Turning angle.}}
        \label{fig:turning_angle}
	\end{subfigure}
	\begin{subfigure}[t]{0.30\linewidth}
		\centering
        \includegraphics[width=1.0\linewidth]{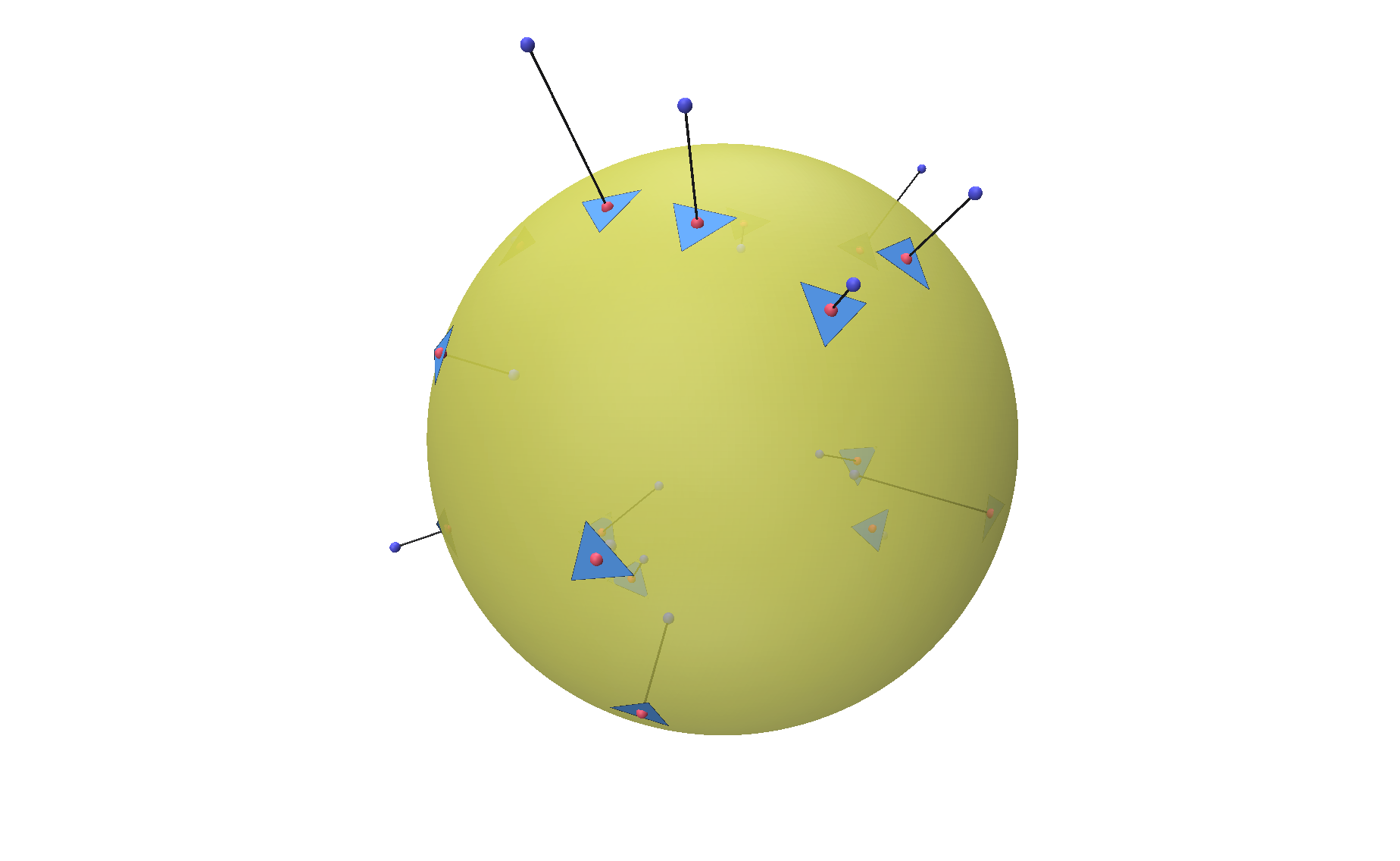}
        \caption{\textbf{Initialization.}}
        \label{fig:mesh_init_triangles}
	\end{subfigure}
	\begin{subfigure}[t]{0.43\linewidth}
		\centering
        \includegraphics[width=1.0\linewidth]{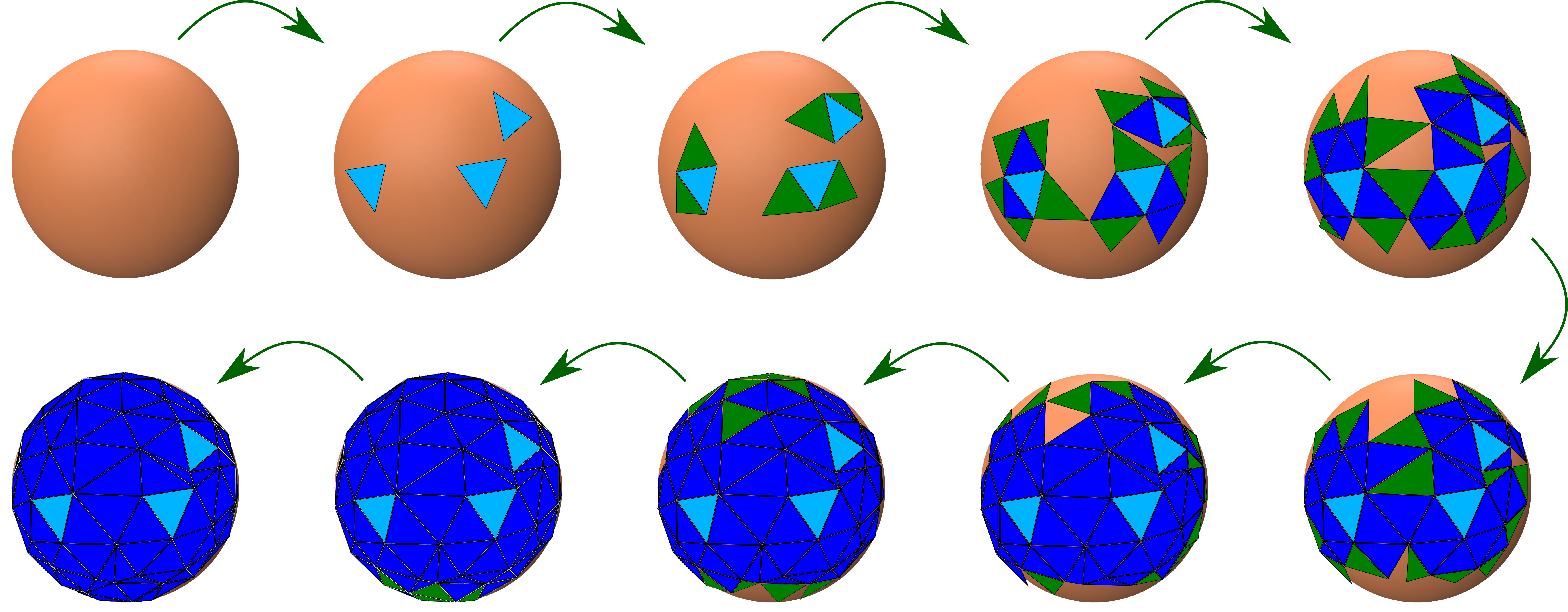}
        \caption{\textbf{Iterative meshing process.}}
        \label{fig:mesh_evolution}
	\end{subfigure}
	\caption{\textbf{Meshing procedure.} 
	(a) We employ the turning angle for approximating the curvature, \ie, the signed angle between the tangential lines, defined by query point $\Xq_1$ and the target point $\Xq_2$, respectively.
	(b) \shortTitle{} first randomly places points in space (blue points), which are then projected to the zero surface of $\XNIP$. The resulting projections (red points) serve as the initialization locations for new surface triangles. 
	(c) Based on the implicit shape (orange sphere), a set of initialization triangles (light blue) is placed on the surface of the object. In an iterative fashion, new faces (green triangles) are added at boundary edges of the existing faces (blue triangles) until the mesh is complete
	}
\end{figure}
\boldparagraph{Curvature Information.}
Our method grows the mesh seeds by generating new triangles in orthogonal direction of unprocessed boundary edges.
For generating new vertices close to the implicit surface, we are interested in the curvature only along that particular direction, \ie, normal curvature, which we found to provide more meaningful information than aggregated curvature measures, \eg, mean or gauss curvature.
The problem of measuring normal curvature on a surface in 3D can be reduced to a line on a 2D-plane, defined by a query point $\Xq_1$, its corresponding surface normal $\Xn_1$ and a query direction vector $\Xqdir$.
To this end, we employ the turning angle for approximating the curvature, \ie, the signed angle between the tangential lines, defined by query point $\Xq_1$ and the target point $\Xq_2$, respectively.
For our case, $\Xq_2$ is computed by following the geodesic path on the discrete mesh along $\Xqdir$ until a fixed distance $\Xldist$ has been covered, as illustrated in Figure~\ref{fig:turning_angle}. 
We define $\Xangle_1=\frac{\pi}{2}$ as the angle between the tangential line and the surface normal at $\Xq_1$, and $\Xangle_2$ as the angle between the tangential line at $\Xq_2$ and the surface normal $\Xn_1$ at $\Xp_1$. 
The turning angle is then computed as
$\Xcurv_{\Xq_1,\Xqdir} = \Xangle_2 - \Xangle_1 = \Xangle_2 - \pi / {2}$ .
The resulting value is positive for surfaces bending away from the surface normal $\Xn_1$, negative for surfaces bending towards the surface normal $\Xn_1$ and zero for flat surfaces.
Furthermore, the distance $\Xldist$ determines the scale of detected curvatures and is fixed to $0.005$ for all of our experiments. 
Although just an approximation of curvature, we found the turning angle to be a good indicator for predicting the amount of surface bending.

\boldparagraph{Directed Curvature Head.}
We extend the neural implicit representation of IGR~\cite{gropp_implicit_2020} with an additional directed curvature head in order to predict the curvature $\Xcurv_{\Xq,\Xqdir}$ 
of a surface point $\Xq$ along the tangential direction $\Xqdir$. 
The extended signed distance function 
$(\Xsdf_{\Xq}, \Xcurv_{\Xq,\Xqdir}) = \Xnet_\Xcurv((\Xq, \Xqdir); \Xnetw; \Xlat)$
returns a tuple of signed distance and normal curvature at point $\Xq$. 
We make the curvature query optional, such that the extended decoder can be queried for signed distance values only.
We use the same training losses as in IGR~\cite{gropp_implicit_2020}.
We train the curvature head with a supervised $L_2$ loss on top of the existing IGR losses $\Xloss_{\text{IGR}}$:
\begin{equation}
\begin{aligned}
    \Xlosstot(\Xnetw; \Xlat) = \Xlossigr(\Xnetw; \Xlat) &+ \Xlosswcurv \mathbb{E}_{\Xq,\Xqdir}(\norm{\Xcurv_{\Xq,\Xqdir} - \Xgtcurv}_2^2), 
\end{aligned}
\end{equation}
where $\Xcurv_{\Xq,\Xqdir}$ is the curvature prediction and $\Xcurv_{\text{GT}}$ the ground truth curvature, based on the turning angle approximation. 
During training, we sample a surface-tangential direction $\Xqdir$ uniformly at random for every element in the batch.
We refer to the supplementary material for architectural details.

\subsection{Iterative Meshing Procedure}
The input to our \textit{iterative meshing module} are the predictions of our modified implicit representation $\XNIP$.
The resolution of the output mesh $\XM = (\XV, \XF)$ is defined by the default equilateral triangle with circumradius $\Xrd$ which is provided as an input parameter.
In an initial step, random faces are placed along the surface defined by $\XNIP$.
Further processing then iteratively extends existing triangles by inserting new faces along boundary edges until completion of the mesh, as shown in Figure~\ref{fig:mesh_evolution}.
In order to keep track of boundary edges, we employ a halfedge data structure, which provides efficient operations for boundary edge access, vertex insertion and face insertion.
To find vertices in a local region quickly, 
we use an additional k-d tree to keep track of mesh vertices.

\boldparagraph{Initialization.}
As visualized in Figure~\ref{fig:mesh_init_triangles}, the set of initial triangles $\{\Xtri_i\}^{\Xnumtriangles}$ is computed by first sampling a set of \Xnumtriangles{} points uniformly at random within the bounding box of the point cloud $\XPC$.
This set of points is then projected onto the surface using the gradient $ \nabla_{\Xq}\Xnet(\Xq;\Xlat)$ 
of the SDF prediction. For each projected point, we construct an equilateral triangle on the tangential plane with circumradius $\Xrd$ and random location of vertices.
We enforce a minimum euclidean distances $\Xdist_{\min} = 3\Xrd$ to be present between all 
projected points in order to avoid overlapping triangles.
We use a k-d tree data structure to query candidates within a radius $\Xdist_{\min}$ efficiently and filter out overlapping triangles.
As the underlying implicit representation $\XNIP$ might exhibit inaccuracies, \eg, far from the surface, we employ a simple heuristic in order to accurately place the initialization triangles.
We perform the surface projection $\Xnumprojections$ times for $k \times \Xnumtriangles$ points, each time bringing the initial random samples closer to the surface, similar to Chibane~\etal~\cite{chibane_neural_2020}.
Finally, we choose random $\Xnumtriangles$ non-overlapping points as initialization locations for the triangles.

\boldparagraph{Iterative Face Insertion.}
Given the initial set of triangles $\{\Xtri_i\}^{\Xnumtriangles}$, our iterative meshing module proceeds to iteratively select boundary edges, \ie, edges of existing triangles which only have one connected face, and predict new vertices until no boundary edges are left.  
The process can be accelerated by computing batches of vertex insertions.
To this end, the vertex predictions are computed batchwise on the GPU.
As batch processing introduces the risk of inserting overlapping triangles, we employ two simple strategies: 
\textbf{(1)} Only boundary edges with distances between their mid-points greater than a certain threshold are processed in the same batch and
\textbf{(2)} we filter overlapping face insertion candidates before adding the faces to the mesh.
This procedure ensures that no overlapping triangles are inserted, given a reasonable threshold.
Since the sampling of non-overlapping boundary edges reduces the number of available boundary edges, we use a simplified procedure in practice where we apply a minimum threshold of $3\Xrd$ between the boundary edge centers.
Although this reduces the number of collisions, the absence of overlaps is not guaranteed.
We apply a final overlap check in order to reject candidates with small distances between triangle centers. 

\subsection{Merging Surface Patches}
Naively inserting new faces for every boundary edge leads to overlapping surface patches.
Therefore, we employ a deterministic procedure for merging such faces.
Prior to creating the triangle proposed by the vertex prediction, we replace it by the existing vertex closest to the center $\Xbecent$ of the boundary edge, 
provided the vertical distance is below a threshold $t_v=\frac{r_d}{2}$
and the prospective new triangle does not overlap with an existing triangle.
\begin{figure}[tb]
    \captionsetup[subfigure]{aboveskip=2pt,belowskip=-5pt}
	\centering
	\begin{subfigure}[t]{0.16\linewidth}
		\centering
		\includegraphics[width=1.\linewidth]{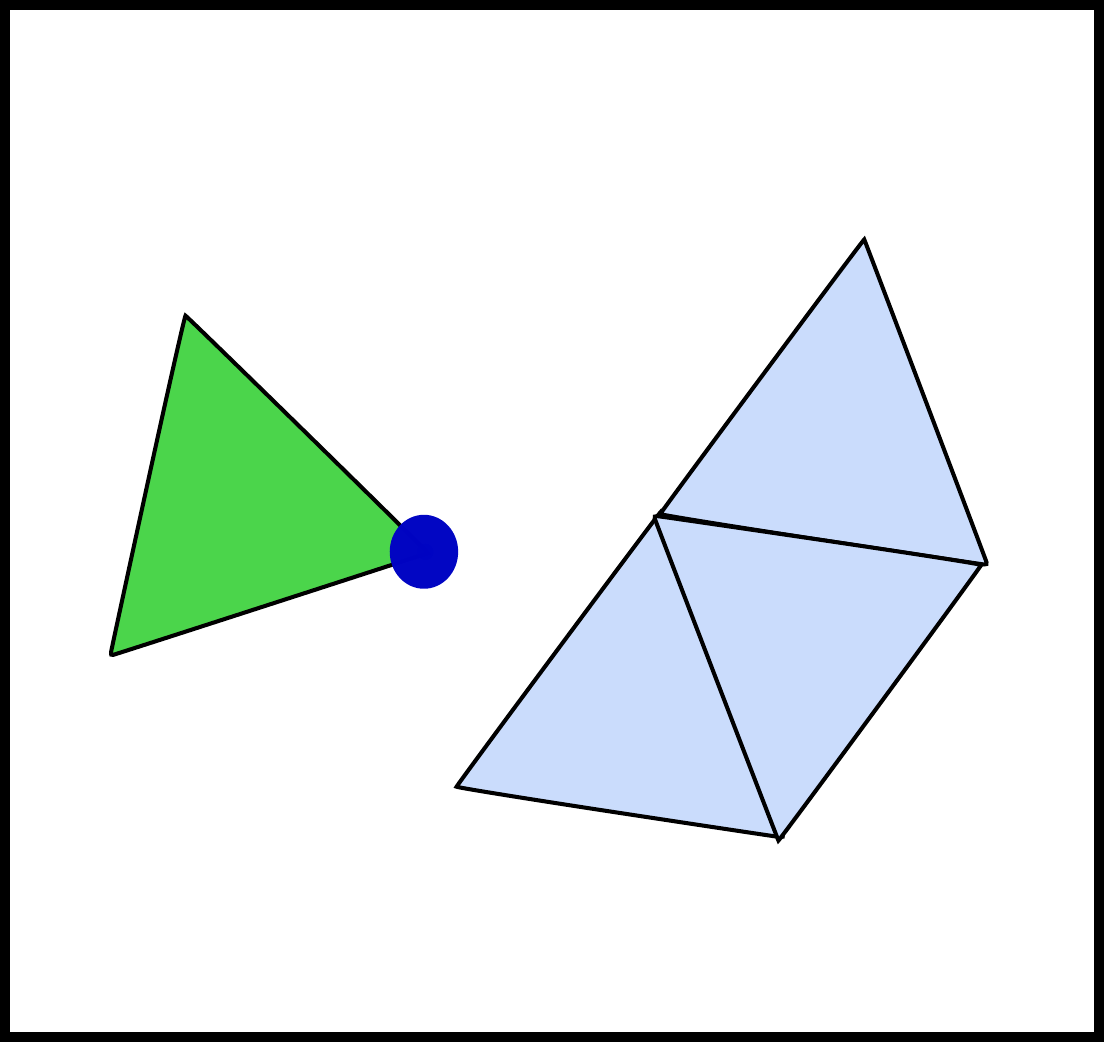}
		\caption{}
		\label{fig:overlap_nope}
	\end{subfigure}
	\begin{subfigure}[t]{0.16\linewidth}
		\centering
		\includegraphics[width=1.\linewidth]{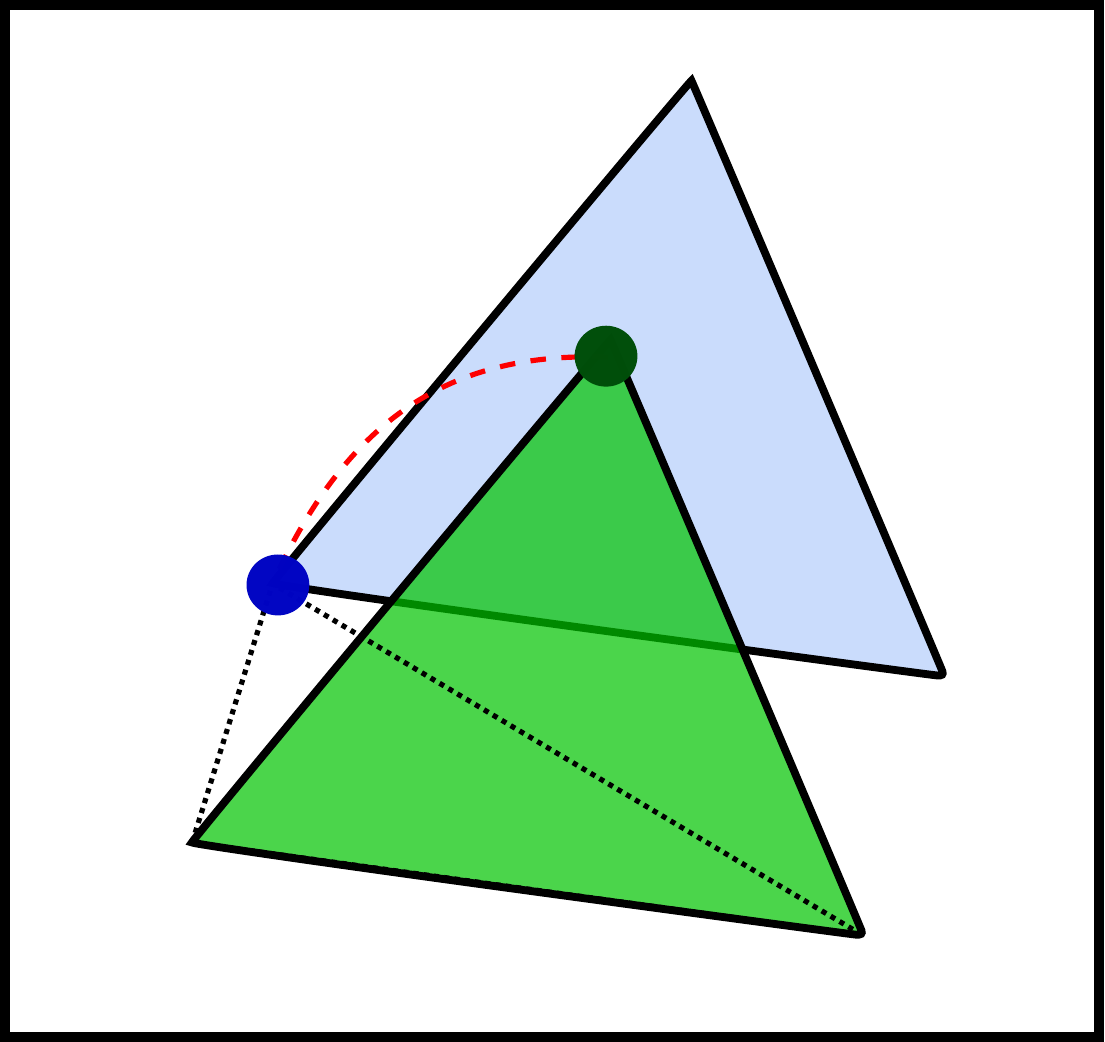}
		\caption{}
		\label{fig:overlap_trivial}
	\end{subfigure}
	\begin{subfigure}[t]{0.16\linewidth}
		\centering
		\includegraphics[width=1.\linewidth]{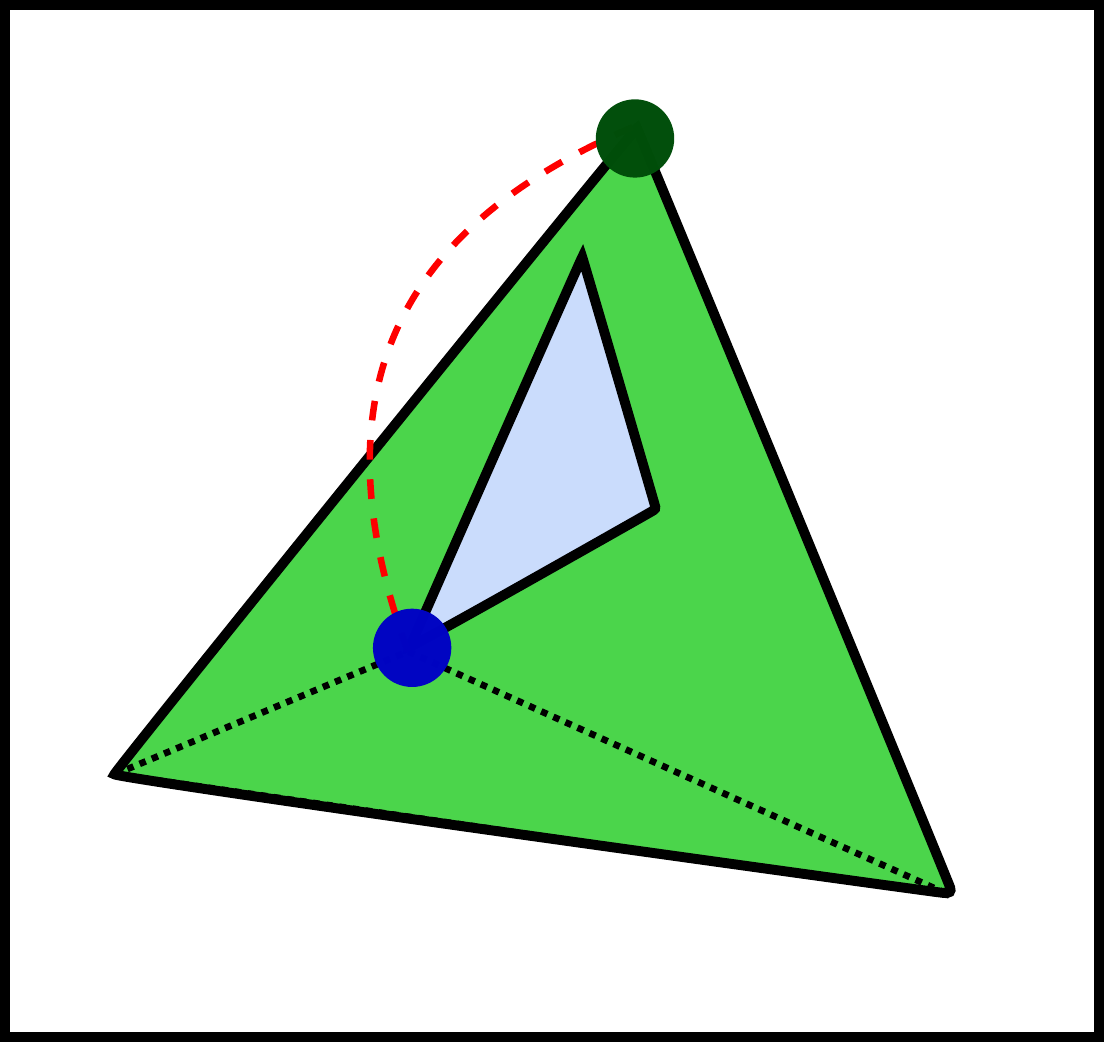}
		\caption{}
		\label{fig:overlap_covering}
	\end{subfigure}
	\begin{subfigure}[t]{0.16\linewidth}
		\centering
		\includegraphics[width=1.\linewidth]{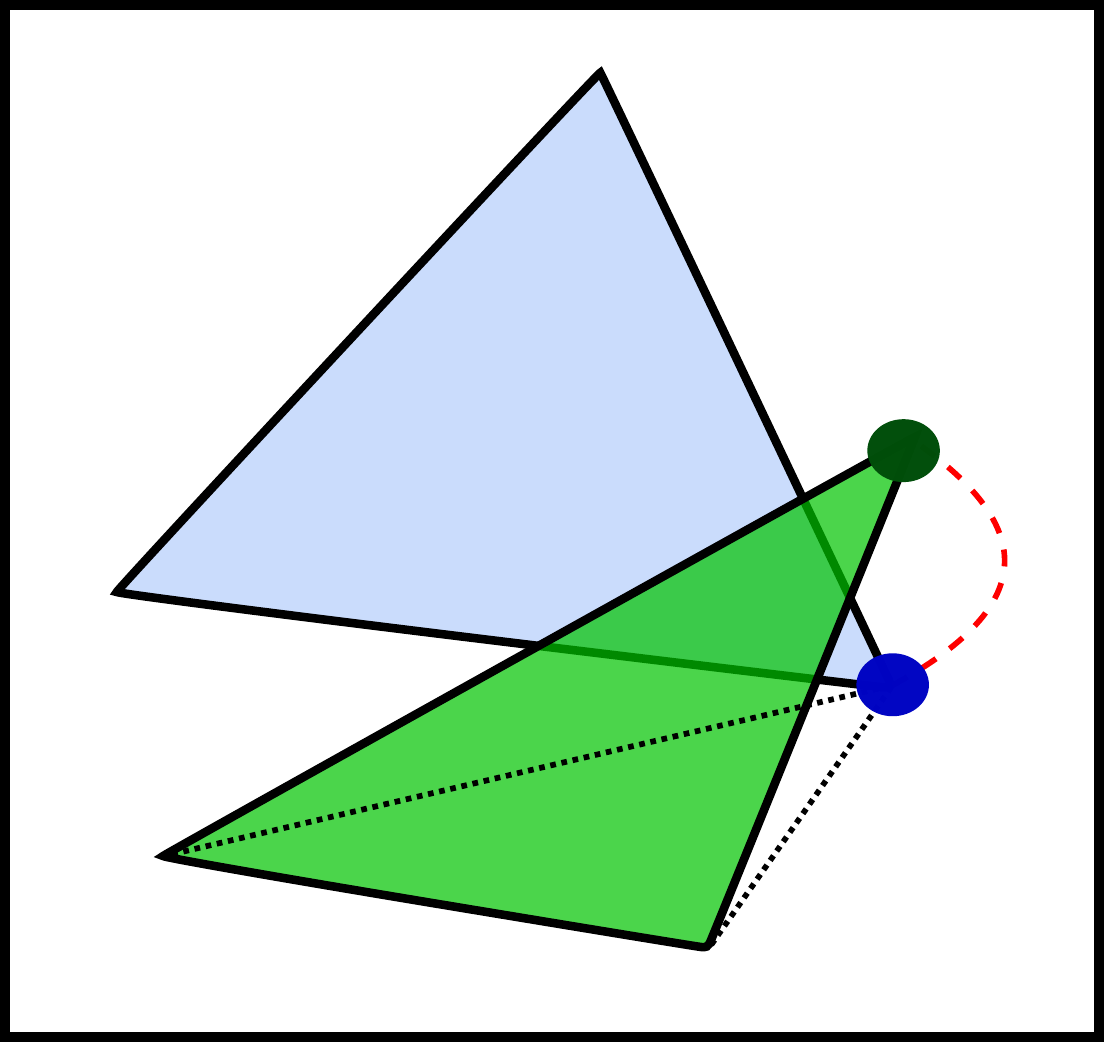}
		\caption{}
		\label{fig:overlap_edge_intersection}
	\end{subfigure}
	\begin{subfigure}[t]{0.16\linewidth}
		\centering
		\includegraphics[width=1.\linewidth]{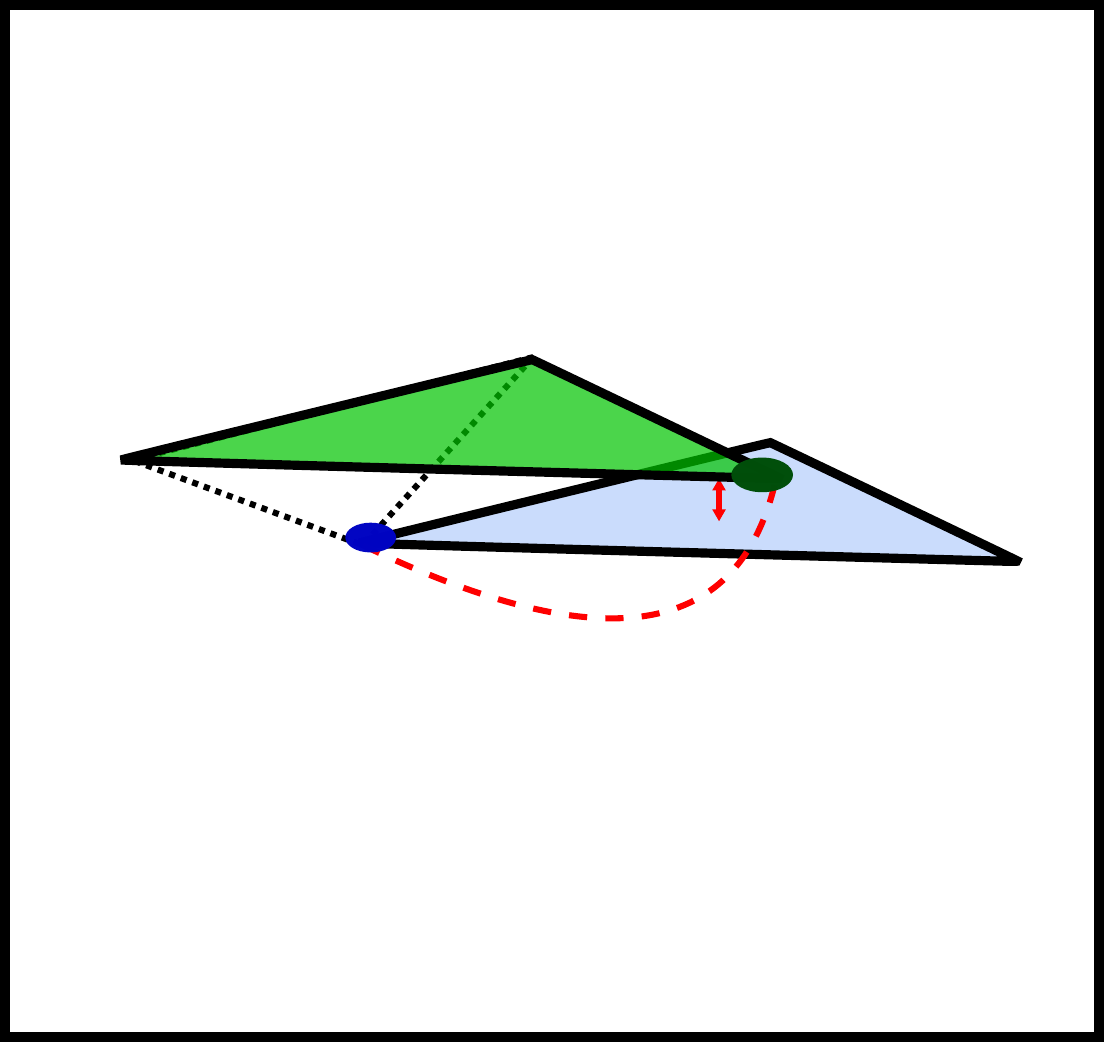}
		\caption{} 
		\label{fig:overlap_parallel}
	\end{subfigure}
	\begin{subfigure}[t]{0.16\linewidth}
		\centering
		\includegraphics[width=1.\linewidth]{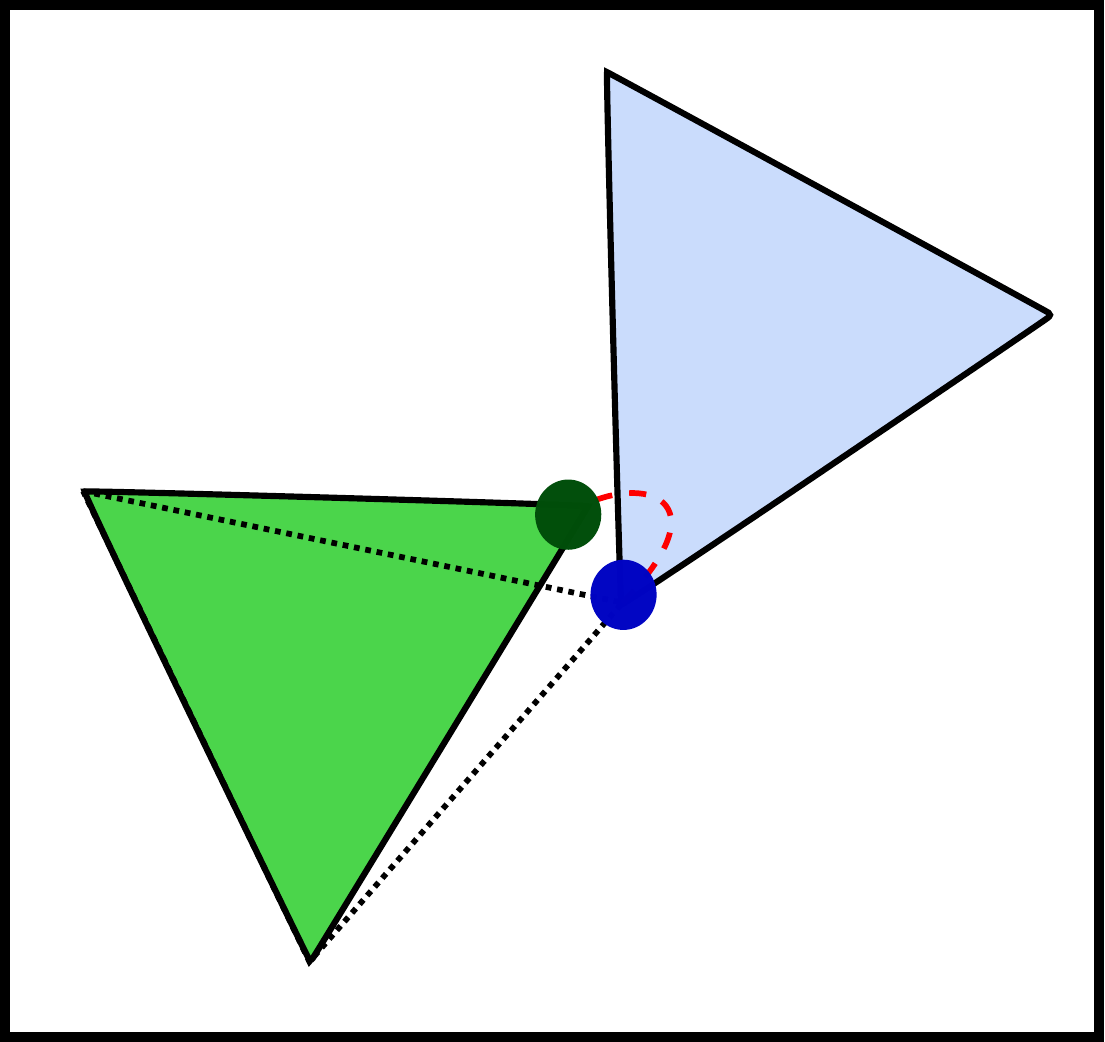}
		\caption{}
		\label{fig:overlap_near_clip}
	\end{subfigure}
	   \caption{\textbf{Triangle-overlap scenarios during merging.} Prior to inserting a prospective new triangle (green), a series of checks is performed, measuring the overlap with existing triangles (blue). In overlap cases, the prospective vertex (green) is replaced with the existing vertex (blue). 
	   (a) No overlap. (b) Predicted vertex inside existing triangle. (c) Existing triangle inside predicted triangle. (d) Triangle edge overlap. (e) No overlap, but close vertical proximity of triangles. (f) No overlap, but close proximity of vertices
	   }
	   \label{fig:overlap}
\end{figure}
In order to decide whether two triangles overlap, we distinguish between several scenarios illustrated in Figure~\ref{fig:overlap}.
For the most frequent case \ref{fig:overlap_nope} where no overlap occurs, the prediction is not replaced.
Case~\ref{fig:overlap_trivial} handles 
geometric intersections between 
the predicted vertex and the existing triangle, while \ref{fig:overlap_covering} addresses the case where an existing triangle is completely contained in the prospective new triangle. 
In both cases, the predicted vertex is projected onto the triangle plane followed by a simple inside-outside test.
Since vertex projection is not sufficient in all cases, we additionally perform edge intersection tests (\ref{fig:overlap_edge_intersection}) and compare the vertical distance between triangle planes (\ref{fig:overlap_parallel}).
Finally, we consider vertices in close proximity to the prediction to be overlapping, as shown in Figure~\ref{fig:overlap_near_clip}.
Please refer to the supplementary material for more details.

\subsection{Vertex Prediction}

\begin{figure}[tb]
    \captionsetup[subfigure]{aboveskip=1pt,belowskip=-5pt}
	\centering
	\begin{subfigure}[t]{0.19\linewidth}
		\centering
		\includegraphics[width=\linewidth]{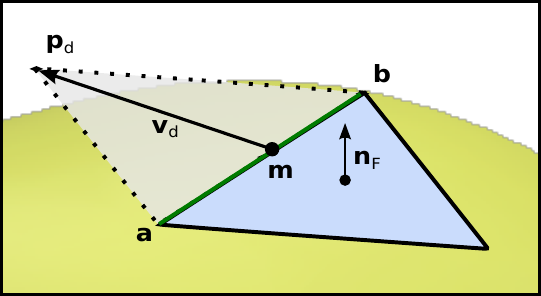}
		\caption{}
		\label{fig:face_insertion_default}
	\end{subfigure}
	\begin{subfigure}[t]{0.19\linewidth}
		\centering
		\includegraphics[width=\linewidth]{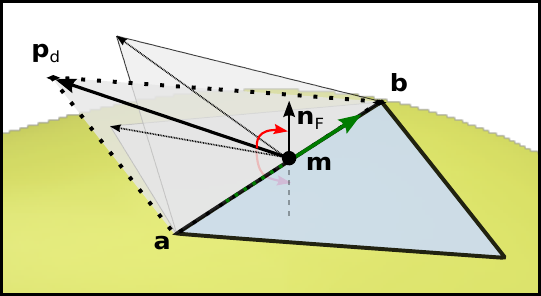}
		\caption{}
		\label{fig:face_insertion_boundary_edge_rot}
	\end{subfigure}
	\begin{subfigure}[t]{0.19\linewidth}
		\centering
		\includegraphics[width=\linewidth]{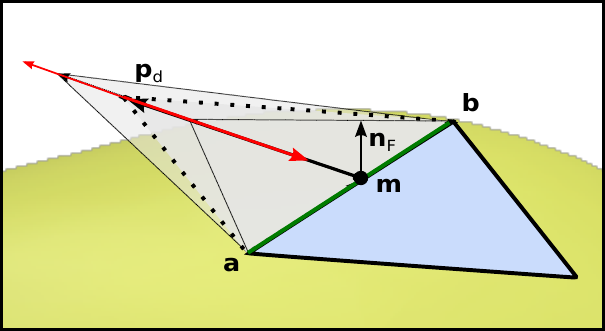}
		\caption{}
		\label{fig:face_insertion_lenth_scaling}
	\end{subfigure}
	\begin{subfigure}[t]{0.19\linewidth}
	    \centering
        \includegraphics[width=1.0\linewidth, clip,trim={0em 0em 19em 0em}]{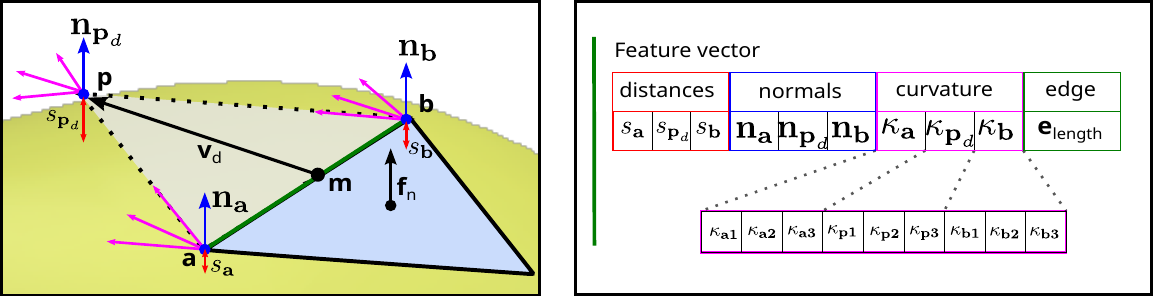}
        \caption{}
        \label{fig:feature_vector_a}
    \end{subfigure}
    \begin{subfigure}[t]{0.19\linewidth}
	    \centering
        \includegraphics[width=1.0\linewidth, clip,trim={18em 0em 0em 0em}]{fig_feature_vector.pdf}
        \caption{}
        \label{fig:feature_vector_b}
    \end{subfigure}
        \caption{\textbf{Vertex prediction and Feature embedding.} In order to predict vertex $\Xpredvert$, we parameterize the prediction with 2 of the available 3 degrees of freedom, \ie, the angle $\scalaredgerotation$ around the boundary edge (b) and the height $\scalarlengthscaling$ of the prospective triangle (c). Note that we omit the angle around the face normal of the triangle adjacent to the boundary edge (green) for practical reasons. By default, the predictor returns the default vector $\defaultprediction$, extending the surface along the same plane as defined by the boundary edge and the corresponding triangle, shown in (a).
        Input to the predictor is a $22$-dimensional feature vector conditioned on the given boundary edge (d+e). 
        It contains the SDF values, the gradients of the SDF values and the curvature predictions from our modified implicit representation $\XNIP$ at the three vertices of the default triangle, \ie, boundary edge end points $\Xbestart$ and $\Xbeend$, and 
        the vertex at the default location $\defaultprediction$. Best viewed digitally
	   }
	   \label{fig:face_insertion_manipulation}
\end{figure}

We introduce a novel \textit{vertex prediction module} which takes a boundary edge and a feature vector as input and predicts the location of the next vertex based on local geometry information.
The output of the predictor is a 2-dimensional vector $\Xpred = 
\begin{bmatrix} 
\scalarlengthscaling & \scalaredgerotation %
\end{bmatrix}$.
As demonstrated in Figure~\ref{fig:face_insertion_manipulation}, both target a separate degree of freedom for
transforming the default prediction $\defaultprediction$ at the center of the boundary edge $\Xbecent$. 
Note that we define $\defaultprediction$ such that it is orthogonal to the boundary edge and the face normal $\Xfnormal$ and incidentally defines the height of the predicted triangle. 
The first component denotes the \textit{boundary edge rotation},
\ie the angle $\scalaredgerotation \in \left[-\pihalbe, \pihalbe \right] $ defining the rotation around the boundary edge (Figure~\ref{fig:face_insertion_boundary_edge_rot}). 
The \textit{length scaling} factor $\scalarlengthscaling \in \left[-1,1\right]$ scales the length of the default vector $\defaultprediction$.
Negative values decrease the vector length whereas positive values increase its length.
Note that we intentionally only predict two out of 3 DoF, 
\ie, we omit the rotation around the face normal $\Xfnormal$ belonging to the face of the boundary edge,
since we
found no performance improvement in incorporating the \nth{3} DoF.

\boldparagraph{Feature Embedding.}
To effectively predict accurate vertex locations, we provide the predictor with a feature embedding containing information about the local geometry.
Figure~\ref{fig:feature_vector_b} depicts the used feature vector.
We consider the 3 points defined by the default triangle, \ie, the vertex at the default location $\defaultprediction$ and the two boundary edge end points, $\Xbestart$ and $\Xbeend$.
For each of these 3 points, we query the implicit representation $\XNIP$ for SDF values ($\XbeSDFc$, $\XbeSDFs$ and $\XbeSDFe$), SDF gradients or normals ($\XbeGradc$, $\XbeGrads$ and $\XbeGrade$) and directional curvature values ($\XbeCurvc$, $\XbeCurvs$ and $\XbeCurve$) (Figure~\ref{fig:feature_vector_a}).
In practice, we use multiple directional curvature queries for each query point.
Additionally, we feed the length of the boundary edge $e_{\text{length}} = \norm{\Xbestart - \Xbeend}_{2}$ into the feature vector.
Note that the feature embedding can be made invariant to rotation and translation. Prior to inference, we therefore transform the normals into a local coordinate system with the boundary edge center $\Xbecent$ as its origin. 
The network predictions can be readily applied within the world coordinate system to obtain the new vertex $\Xpredvert$.
In order to reduce prediction errors, we apply surface projection once as a post-processing step in the same way as performed during the initialization phase.

\boldparagraph{Loss Functions.}
We introduce a surface distance loss $\Xlosssd=\abs{\Xpredsdf}$, in order to penalize any deviation from the zero value for the SDF $\Xpredsdf$ of the predicted vertex $\Xpredvert$. 
To encourage the network to predict triangles with default size $\Xrd$, we additionally define a length regularization loss $\Xlossreg=\abs{ \scalarlengthscaling }$ which prevents prediction of degenerate triangles close to the boundary edge and competes with the surface distance loss.
Based on the surface mapping procedure, applied as a post-processing step, we can compute the ground truth turning angle $\Xgtangle$, located at the boundary edge. 
Therefore, the boundary edge rotation loss $\Xlossbe= \abs{\scalaredgerotation - \Xgtangle}$ encourages the network to predict vertices close to the surface and penalize the predicted boundary angle $\scalaredgerotation$.
The final loss then consists of a weighted sum of those loss terms, \ie,
$
\Xloss_{\text{Total}} = \Xlosswsd  \Xlosssd + \Xlosswbe  \Xlossbe + \Xlosswreg  \Xlossreg.
$

%% file: 04_evaluation.tex
\section{Evaluation}
\label{sec:evaluation}
For the experiments, we use a subset of the D-Faust~\cite{bogo_dynamic_2017} dataset, containing high-resolution scans of humans in different poses and corresponding triangle meshes, which we use as ground truth (GT) reference.
We train, validate and test on 512, 64 and 32 poses respectively,
sampled randomly from the subset used in IGR~\cite{gropp_implicit_2020}.
To further evaluate our capability of dealing with sharp corners and edges, we evaluate on a selected subset of shapes, belonging to the \textit{file cabinet} category of the ShapeNet~\cite{chang_shapenet_2015} dataset.
We use 147 models for training, 32 for validation and 32 for testing and preprocess the models with ManifoldPlus~\cite{huang2020manifoldplus}.

\subsection{Reconstruction Quality} 
\input{tab_chamfer_distances}
\boldparagraph{Reconstruction Error.}
We evaluate the reconstruction error of the produced triangle meshes with the Chamfer-L1 distance 
in Table~\ref{tab:eval_chamfer_distances}. 
The Chamfer distance is reported in two directions, \ie, from the prediction to the ground truth and the implicit representation, respectively, and vice versa.
We report both, the distance to the ground truth mesh and the distance to the respective implicit representation, since there is a discrepancy between both, introducing an additional error in the reported numbers. 
We compare our method to the SotA method PointTriNet~\cite{sharp_pointtrinet_2020}, however, since PointTriNet directly operates on point clouds sampled from the ground truth mesh, the reported error appears lower than methods working directly on implicit representations.
We therefore evaluate PointTriNet on both, points sampled on the ground truth mesh, and points generated from the implicit representation by first sampling surface points on the ground truth mesh and projecting them to the zero level set of the implicit representation.
We further compare our method to a version of marching cubes~\cite{lewiner_efficient_2003} implemented by scikit-image~\cite{scikit-image}, which we evaluate on three different resolutions.
In the same manner, we evaluate our method for three different triangle sizes $\Xrd$.
Finally, we compare to PSR~\cite{kazhdan_poisson_2006} from Open3D (depth=$10$), and DSE~\cite{rakotosaona_learning_2021}, with the author-provided model.
On the highest resolution, our method outperforms all baselines, when measured on the implicit representation, while yielding comparable results on the medium resolution. 
Compared to marching cubes, our method improves on every comparable level of resolution.
For evaluations on the ground truth mesh, \shortTitle{} also outperforms marching cubes.
PointTriNet evaluated on the ground truth sampled points shows comparatively better numbers, which is expected because of the error introduced by the underlying implicit representation.

\boldparagraph{Reconstruction Accuracy and Completeness.}
\input{tab_f1_score_both_all}
We report further quantitative results in Table~\ref{tab:results_all_f1score_both}.
Specifically, we report the accuracy, completeness and the F-score for 3 different inlier thresholds, evaluated on both implicit surface representation and ground truth mesh. 
The accuracy is computed as the ratio of inlier points, \ie sampled points on the predicted mesh which are within inlier distance of the ground truth mesh, and total number of sampled points.
The completeness is computed similarly in the opposite direction.
Our method again outperforms marching cubes on implicit surface evaluations while PointTriNet performs better when evaluated on the ground truth mesh, presumably because of reconstruction errors inherent in the implicit surface representation. 

\boldparagraph{Qualitative Results.} 
\begin{figure}[tb]
    \includegraphics[width=1.\linewidth, clip,trim={0em 0em 0em 0em}]{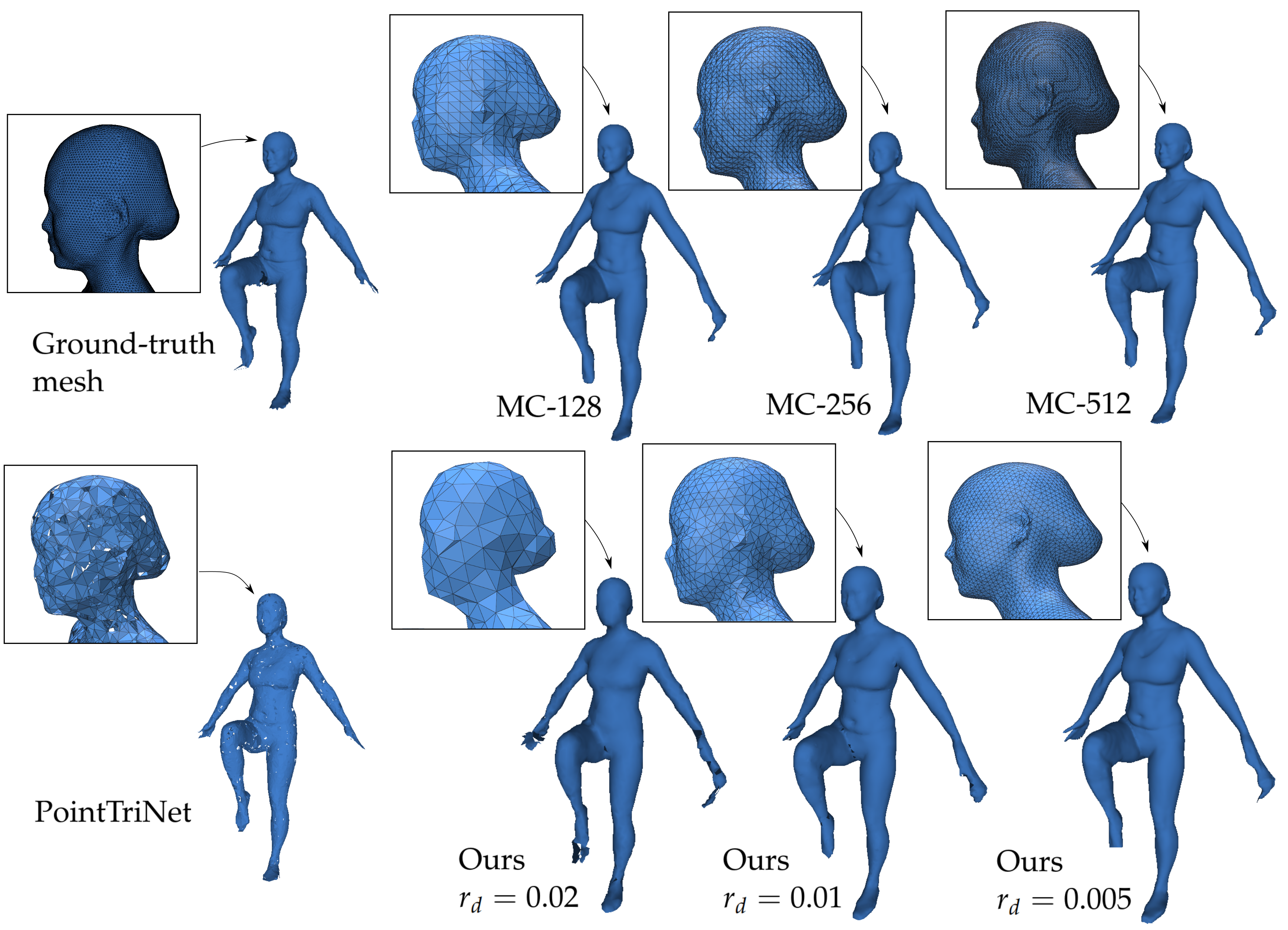}
    \caption{\textbf{Qualitative results.} Result meshes of various baselines from a single sample. The level of detail reconstructed on the face heavily depends on the resolution and method. IGR artifacts were manually removed. Best viewed digitally}
    \label{fig:meshes_2571}
\end{figure}
Figure~\ref{fig:meshes_2571} shows the meshes of all methods on one example.  
\shortTitle{} yields well behaved triangle meshes with regularly shaped and sized triangles, reconstructing high-level details in accordance to the chosen triangle size $\Xrd$.
While marching cubes on comparable resolution levels reconstructs a similar level of detail, their generated triangle sizes cannot become bigger than the underlying voxel size.
PointTriNet reconstructs the full shape with some level of detail but introduces many visible holes and overlapping faces.

\subsection{Triangle Mesh Properties}
In this section, we evaluate several mesh properties and triangle metrics.
\boldparagraph{Mesh Metrics.}
\input{tab_mesh_sizes}%
To demonstrate the capability of producing detail-preserving, low-memory triangle meshes, we report typical mesh metrics in Table~\ref{tab:mesh_sizes}, \ie, number of triangles, number of vertices and the average triangle area. 
We also list the inference time for speed comparisons.
Our method produces less but bigger triangles than marching cubes for comparable levels of detail. 

\boldparagraph{Distribution of Face Area, Triangle Angles and Holes.}
\begin{figure}[tb]
	\centering
    \captionsetup[subfigure]{aboveskip=0pt,belowskip=-10pt}
	\begin{subfigure}[t]{0.32\linewidth}
        \centering
		\includegraphics[width=\linewidth]{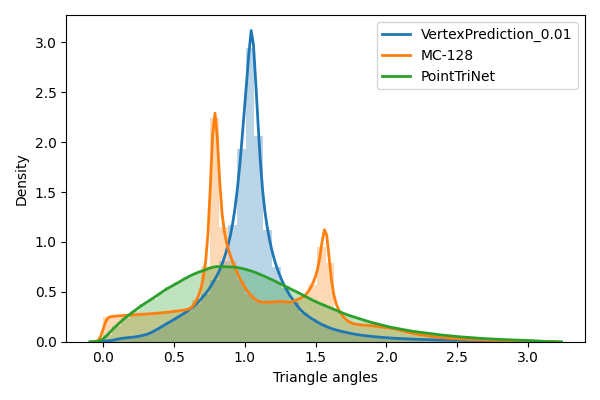}
		\caption{}
		\label{fig:eval_angle_histogram}
	\end{subfigure}
	\begin{subfigure}[t]{0.32\linewidth}
		\centering
        \includegraphics[width=\linewidth]{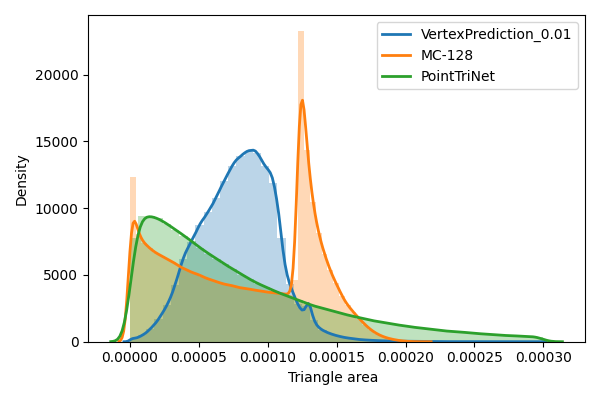}
		\caption{}
        \label{fig:eval_area_histogram}
	\end{subfigure}
	\begin{subfigure}[t]{0.32\linewidth}
		\centering
        \includegraphics[width=\linewidth]{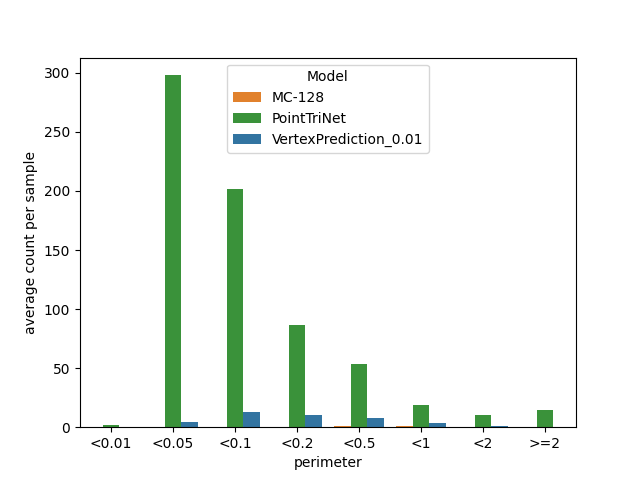}
		\caption{}
        \label{fig:hole_dist}
	\end{subfigure}
	\caption{\textbf{Distribution analysis.} We compare the distribution of triangle angles in radians (a), triangle areas (b) and average number of holes (c) between our method, PointTriNet~\cite{sharp_pointtrinet_2020} and marching cubes~\cite{lewiner_efficient_2003}
	on the D-Faust~\cite{bogo_dynamic_2017} dataset
	}
\end{figure}
To demonstrate the regularity of the generated faces, Figure~\ref{fig:eval_angle_histogram} shows the angle distribution observed in generated meshes of each method.
Compared to marching cubes and PointTriNet, 
our method produces triangle angles closer to the equilateral triangle, containing very few triangles with tiny angles. 
Figure~\ref{fig:eval_area_histogram} provides similar insights by comparing the triangle area distribution. 
It can be observed that our method produces more similarly sized faces while still allowing some variation in order to adapt to more complex surface patches.
In Figure~\ref{fig:hole_dist}, we plot the number of holes vs. the hole size and compare it with both baselines, illustrating that the vast majority of holes produced by \shortTitle{} are very small.
We refer to the supplementary material for more quantitative metrics and ablation studies.

\boldparagraph{Limitations.}
\shortTitle{} does not guarantee watertightness, but typically produces meshes with fewer holes than comparable methods.
Sharp edges are sometimes problematic which we attribute to the performance of the prediction network.
Like marching cubes, \shortTitle{} does not provide a manifoldness guarantee.
However, our mesh growing strategy can effectively avoid the insertion of non-manifold edges, while marching cubes requires expensive post-processing.

%% file: tab_chamfer_distances.tex
\begin{table}[tb]
    \centering
    \caption{\textbf{Reconstruction error on D-Faust~\cite{bogo_dynamic_2017}.} We report the distances of closest point pairs between the generated meshes and the ground truth (GT) or the implicit representation (IGR) respectively. The distances are evaluated at 20k randomly sampled surface points. 
    For the implicit representation the sampled GT points are projected to the zero level set of the implicit representation
    }
    \label{tab:eval_chamfer_distances}
    \setlength{\tabcolsep}{3.5pt} %
    \renewcommand{\arraystretch}{1.05} %
    \resizebox{1.0\linewidth}{!}{
    \begin{tabular}{llrrrrrr}
    \multicolumn{2}{r}{from} & \multicolumn{2}{c}{Generated mesh} & GT & IGR & \multicolumn{2}{c}{Bidirectional} \\
    \cmidrule(lr){3-4}\cmidrule(lr){5-6}\cmidrule(lr){7-8}
    \multicolumn{2}{r}{to} & GT\gb{[1e-4]}$\downarrow$ & IGR\gb{[1e-4]}$\downarrow$ & \multicolumn{2}{c}{Generated Mesh\gb{[1e-4]}$\downarrow$ } & GT\gb{[1e-4]}$\downarrow$ & IGR\gb{[1e-4 ]}$\downarrow$ \\
    \midrule\multirow{3}{*}{ Ours }
    & $\defaultradius=0.02$ & 83.221  & 14.691  & 25.434  &  16.405 & 54.328  & 15.548  \\
    & $\defaultradius=0.01$ & 58.686  & 4.950  & 20.859  &  8.264 & 39.772  & 6.607  \\
    & $\defaultradius=0.005$ & 44.768  & \B 1.096  & 19.994  &  \B 5.643 & 32.381  & \B 3.370  \\
    
    \midrule
    \multirow{3}{*}{ MC \cite{lewiner_efficient_2003} }
    & $res=128$ & 90.298  & 68.332  & 78.490  &  76.412 & 84.394  & 72.372  \\
    & $res=256$ & 60.221  & 34.591  & 43.419  &  39.869 & 51.820  & 37.230  \\
    & $res=512$ & 46.340  & 17.380  & 27.767  &  21.997 & 37.053  & 19.689  \\
    
    \midrule
    \multicolumn{2}{l}{ PointTriNet GT~\cite{sharp_pointtrinet_2020}} &\B 6.552  & 16.626  & \B 7.634  &  17.495 & \B 7.093  & 17.061  \\
    \multicolumn{2}{l}{ PointTriNet IGR~\cite{sharp_pointtrinet_2020}} & 38.156  & 5.519  & 23.758  &  12.747 & 30.957  & 9.133  \\
    
    \midrule
    \multicolumn{2}{l}{PSR \cite{kazhdan_poisson_2006}} 	& 38.764 & 2.3667 & 19.086 & 4.444   & 28.925 & 3.405 \\
    \multicolumn{2}{l}{DSE \cite{rakotosaona_learning_2021}} 	& 43.021 & 5.5607 & 22.892 & 11.612  & 32.957 & 8.586 \\
    \end{tabular}
    }
\end{table}

%% file: tab_f1_score_both_all.tex
\begin{table}[tb]
    \centering
    \caption{\textbf{Reconstruction accuracy and completeness on D-Faust~\cite{bogo_dynamic_2017} (top) and ShapeNet~\cite{chang_shapenet_2015} (bottom).} We report accuracy, completeness and F1-score for 3 different inlier thresholds when evaluated on the implicit representation or the ground truth mesh respectively
    }
    \label{tab:results_all_f1score_both}
    \setlength{\tabcolsep}{3.5pt} %
    \renewcommand{\arraystretch}{1.05} %
    \resizebox{1.0\linewidth}{!}{
    \begin{tabular}{llcccccccccHHH}
    \multicolumn{2}{l}{IGR / GT} & \multicolumn{3}{c}{$d_{inlier}=0.001 (0.05\%)$} & \multicolumn{3}{c}{$d_{inlier}=0.005 (0.25\%)$} &  \multicolumn{3}{c}{$d_{inlier}=0.01 (0.5\%)$} & \\% \multicolumn{3}{c}{$d_{inlier}=0.02 (1\%)$} \\
    \cmidrule(lr){3-5}\cmidrule(lr){6-8}\cmidrule(lr){9-11}\cmidrule(lr){12-14}
    \multicolumn{2}{r}{} & Acc. $\uparrow$ & Com. $\uparrow$ & F1 $\uparrow$ & Acc. $\uparrow$ & Com. $\uparrow$ & F1 $\uparrow$ & Acc. $\uparrow$ & Com. $\uparrow$ & F1 $\uparrow$ & Acc. $\uparrow$ & Com. $\uparrow$ & F1 $\uparrow$ \\
    \toprule\multirow{3}{*}{ Ours }
    & $\defaultradius=0.02$ & 0.406 / 0.136   & 0.535 / 0.317 & 0.453 / 0.180
                   & 0.620 / 0.339   & 0.839 / 0.793 & 0.700 / 0.447
                   & 0.660 / 0.369   & 0.863 / 0.851 & 0.732 / 0.484
                   & 0.732 / 0.386   & 0.873 / 0.870 & 0.776 / 0.503 \\
    & $\defaultradius=0.01$ & {\B 0.975} / 0.479   & 0.787 / 0.393 & 0.870 / 0.431
                   & {\B 1.000} / 0.953   & 0.830 / 0.802 & 0.906 / 0.870
                   & {\B 1.000} / 0.982   & 0.851 / 0.845 & 0.919 / 0.907
                   & {\B 1.000} / 0.992   & 0.880 / 0.878 & 0.935 / 0.931 \\
    & $\defaultradius=0.005$ & 0.971 / 0.452   & {\B 0.968} / 0.475 & {\B 0.969} / 0.463
                   & 0.989 / 0.898   & {\B 0.985} /  0.947 & {\B 0.987} / 0.922
                   & 0.999 / 0.932   &  0.988 /  0.979 & 0.993 / 0.954
                   & {\B 1.000 } / 0.951   & 0.992 / 0.990 & 0.996 / 0.969 \\
    \midrule
    \multirow{3}{*}{ MC \cite{lewiner_efficient_2003} }
    & $res=128$ & 0.072 / 0.068   & 0.067 / 0.067 & 0.069 / 0.068
                   & 0.362 / 0.346   & 0.339 / 0.341 & 0.350 / 0.343
                   & 0.763 / 0.724   & 0.720 / 0.713 & 0.741 / 0.718
                   & {\B 1.000} / 0.972   & 0.987 / 0.981 & 0.993 / 0.976 \\
    & $res=256$ & 0.140 / 0.133   & 0.133 / 0.134 & 0.136 / 0.134
                   & 0.756 / 0.689   & 0.723 / 0.694 & 0.739 / 0.691
                   & {\B 1.000} / 0.947   & 0.986 / 0.965 & 0.993 / 0.956
                   & {\B 1.000} / 0.973   & 0.990 / 0.988 & 0.995 / 0.980 \\
    & $res=512$ & 0.279 / 0.262   & 0.268 / 0.266 & 0.274 / 0.264
                   & {\B 1.000} / 0.893   & 0.983 / 0.913 & 0.991 / 0.902
                   & {\B 1.000} / 0.955   & 0.987 / 0.976 & {\B 0.994} / {\B 0.965 }
                   & {\B 1.000} / 0.973   & 0.991 / 0.989 & 0.996 / 0.981 \\
    \midrule
    \multicolumn{2}{l}{ PointTriNet  \cite{sharp_pointtrinet_2020} GT  } & 0.445 / {\B 0.831}   & 0.425 / {\B 0.793} & 0.435 / {\B 0.811}
                   & 0.957 / {\B 0.990}   & 0.947 / {\B 0.982} & 0.952 / {\B 0.986}
                   & 0.990 / {\B 0.997}   & {\B 0.991} / {\B 0.998} & 0.991 / 0.997
                   & 0.998 /  {\B 1.000} & {\B 0.999} /  {\B 1.000} & {\B 0.999} / {\B 1.000} \\
    \multicolumn{2}{l}{ PointTriNet  \cite{sharp_pointtrinet_2020} IGR  } & 0.849 / 0.430   & 0.777 / 0.407 & 0.811 / 0.418
                   & 0.998 / 0.922   & 0.965 / 0.924 & 0.981 / 0.923
                   & {\B 1.000} / 0.959   & 0.984 / 0.973 & 0.992 / 0.966
                   & {\B 1.000} / 0.975   & 0.990 / 0.988 & 0.995 / 0.981 \\
    \end{tabular}
    }
    \resizebox{1.0\linewidth}{!}{
    \begin{tabular}{llcccccccccHHH}
    \multicolumn{2}{l}{IGR / GT} & \multicolumn{3}{c}{$d_{inlier}=0.001 (0.05\%)$} & \multicolumn{3}{c}{$d_{inlier}=0.005 (0.25\%)$} &  \multicolumn{3}{c}{$d_{inlier}=0.01 (0.5\%)$} & \\%\multicolumn{3}{c}{$d_{inlier}=0.02 (1\%)$} \\
    \cmidrule(lr){3-5}\cmidrule(lr){6-8}\cmidrule(lr){9-11}\cmidrule(lr){12-14}
    \multicolumn{2}{r}{} & Acc. $\uparrow$ & Com. $\uparrow$ & F1 $\uparrow$ & Acc. $\uparrow$ & Com. $\uparrow$ & F1 $\uparrow$ & Acc. $\uparrow$ & Com. $\uparrow$ & F1 $\uparrow$ & Acc. $\uparrow$ & Com. $\uparrow$ & F1 $\uparrow$ \\
    \toprule
    \multirow{3}{*}{ Ours }
    & $\defaultradius=0.02$ & 0.848 / 0.630   & 0.807 / 0.613 & 0.826 / 0.620 
                   & 0.902 / 0.778   & 0.850 / 0.758 & 0.874 / 0.766 
                   & 0.936 / 0.815   & 0.871 / 0.792 & 0.901 / 0.801 
                   & 0.972 / 0.894   & 0.882 / 0.863 & 0.925 / 0.877  \\
    & $\defaultradius=0.01$ & 0.982 / 0.706   & 0.852 / 0.620 & 0.911 / 0.659 
                   & 0.999 / 0.855   & 0.876 / 0.755 & 0.933 / 0.801 
                   & {\B 1.000} / 0.874   & 0.884 / 0.775 & 0.937 / 0.820 
                   & {\B 1.000} / 0.963   & 0.893 / 0.865 & 0.943 / 0.911  \\
    & $\defaultradius=0.005$ & {\B 0.981} / 0.719   & {\B 0.881} / 0.656 & {\B 0.927} / 0.685 
                   & 0.998 / 0.857   & {\B 0.903} / 0.783 & {\B 0.947 }/ 0.817 
                   & {\B 1.000} / 0.874   & 0.909 / 0.803 & 0.951 / 0.836 
                   & {\B 1.000} / 0.967   & 0.919 / 0.896 & 0.957 / 0.929  \\
    \midrule
    \multirow{3}{*}{ MC \cite{lewiner_efficient_2003} }
    & $res=128$ & 0.007 / 0.013   & 0.004 / 0.011 & 0.004 / 0.012 
                   & 0.206 / 0.174   & 0.163 / 0.144 & 0.181 / 0.158 
                   & 0.635 / 0.624   & 0.524 / 0.529 & 0.573 / 0.572 
                   & {\B 1.000} / 0.982   & 0.873 / 0.861 & 0.932 / 0.917  \\
    & $res=256$ & 0.008 / 0.013   & 0.005 / 0.010 & 0.006 / 0.011 
                   & 0.633 / 0.576   & 0.538 / 0.495 & 0.581 / 0.532 
                   & {\B 1.000} / 0.922   & 0.872 / 0.806 & 0.931 / 0.859 
                   & {\B 1.000} / 0.980   & 0.883 / 0.867 & 0.937 / 0.920  \\
    & $res=512$ & 0.084 / 0.055   & 0.070 / 0.046 & 0.075 / 0.050 
                   &{\B 1.000} / 0.863   & 0.873 / 0.753 & 0.932 / 0.803 
                   & {\B 1.000} / 0.907   & 0.880 / 0.796 & 0.935 / 0.847 
                   & {\B 1.000} / 0.968   & 0.888 / 0.861 & 0.940 / 0.911  \\
    \midrule
    \multicolumn{2}{l}{ PointTriNet  \cite{sharp_pointtrinet_2020}} GT & 0.667 / {\B 0.928}   & 0.615 / {\B 0.845} & 0.640 / {\B 0.884 }
                   & 0.889 / {\B 0.980}   & 0.859 / {\B 0.935} & 0.873 / {\B  0.957 }
                   & 0.920 /{\B  0.997}   & 0.917 / {\B 0.980} & 0.918 / {\B 0.988} 
                   & 0.998 /{\B  1.000}   & 0.998 / {\B 0.998} & {\B 0.998} / {\B 0.999}  \\
    \multicolumn{2}{l}{ PointTriNet  \cite{sharp_pointtrinet_2020}} IGR & 0.945 / 0.605   & 0.807 / 0.512 & 0.870 / 0.554 
                   & 0.984 / 0.877   & 0.884 / 0.782 & 0.931 / 0.826 
                   & 0.997 / 0.896   & {\B 0.924} / 0.822 & {\B 0.959} / 0.857 
                   & {\B 1.000 }/ {\B 1.000}   &{\B  0.943} / 0.938 & 0.970 / 0.967  \\
    \end{tabular}
    }
\end{table}

%% file: tab_mesh_sizes.tex
\begin{table}[tb]
    \centering
    \scriptsize
    \setlength{\tabcolsep}{1pt} %
    \renewcommand{\arraystretch}{1.5} %
    \floatbox[\capbeside]{table}[0.53\columnwidth]%
    {\caption{\textbf{Mesh metrics.} 
    We report number of faces (\#F), number of vertices (\#V), the average triangle area and the run-time in seconds. Where appropriate, values are scaled for better readability. 
    PointTriNet roughly generated similarily sized meshes as \shortTitle{} with $\defaultradius=0.02$. 
    Likewise, marching cubes with a resolution of $128$ produces comparably sized meshes as ours with $\defaultradius=0.01$
    }
    \label{tab:mesh_sizes}}%
    {\begin{tabular}{llrrHrr}
    \multicolumn{2}{l}{\ } & F\gb{[1e3]} & V\gb{[1e3]} & $\frac{ \#F }{ \#V }$ & Area\gb{[1e6]} & Time\gb{[s]} \\
    \midrule\multirow{3}{*}{ Ours }
    & $\defaultradius=0.02$ & 4.93  & 2.52  & 1.95  &  381.28 & 14.28  \\
    & $\defaultradius=0.01$ & 41.73  & 20.97  & 1.99  &  42.82 & 226.64  \\
    & $\defaultradius=0.005$ & 81.95  & 41.13  & 1.99  &  22.27 & 182.47  \\
    \midrule
    \multirow{3}{*}{ MC \cite{lewiner_efficient_2003} }
    & $res=128$ & 21.15  & 10.60  & 1.99  &  83.59 & 3.42  \\
    & $res=256$ & 86.26  & 43.19  & 2.00  &  20.95 & 23.42  \\
    & $res=512$ & 347.55  & 173.91  & 2.00  &  5.24 & 187.29  \\
    \midrule
    \multicolumn{2}{l}{ PointTriNet GT \cite{sharp_pointtrinet_2020}  } & 18.88  & 10.00  & 1.89  &  88.51 & 1530.35  \\
    \multicolumn{2}{l}{ PointTriNet IGR \cite{sharp_pointtrinet_2020}  } & 18.93  & 10.00  & 1.89  &  90.03 & 1862.27  \\
    \end{tabular}}
\end{table}

%% file: 05_conclusion.tex
\section{Conclusion}
\label{sec:conclusion}

We introduced \shortTitle{}, a novel meshing algorithm for neural implicit representations.
We exploit curvature information learned as part of the neural implicit representation in order to guide the predictions of new triangles.
Our iterative, curvature-based processing of boundary edges allows us to generate triangle sizes in accordance to the underlying curvature, yielding preferable mesh properties. 
Experiments demonstrate that \shortTitle{} outperforms existing meshing algorithms,
producing meshes with lower triangle counts.

%% file: supplementary_content.tex
In this supplementary material, we provide more details on overlap checks (Sec.~\ref{sec:overlap_details}), the curvature head of the modified neural implicit representation (Sec.~\ref{sec:curvature_head_details}) and the vertex prediction module (Sec.~\ref{sec:vertex_predictor_details}).
The data preprocessing used for the experiments is explained in Sec.~\ref{sec:data_preprocessing}.
In Sec.~\ref{sec:quantitative_results} and Sec.~\ref{sec:more_qualitative_results}, we provide additional quantitative and qualitative results, respectively. 
We provide ablation studies in Sec.~\ref{sec:ablation_results} and further quantitative results regarding the watertightness property in Sec~\ref{sec:watertightness}.
Finally, we discuss sharp features and highlight a few more limitations in Sec.~\ref{sec:limitations}.

\section{Details on Overlap Checks} 
\label{sec:overlap_details}
The size of the triangles inserted into the mesh is controlled by the circumradius $\Xrd$ of the default triangle while the height of a triangle is upper bounded by two times the height of the default triangle.  
It is therefore sufficient to apply the overlap check only on a local subset of nearby mesh faces. 
For a triangle candidate $f_c$, we define this set of relevant local faces as
\begin{equation}
	F_{f_c} = \{f \in F \ \lvert \  d(f, c(f_c)) \leq 2  \defaultradius  \},
\end{equation}
where $c(f_c)$ is the centroid of the triangle and $d(f, c(f_c))$ denotes the smallest euclidean distance among the face vertices in $f$ and the triangle center of $f_c$.
$F_{f_c}$ can be efficiently constructed by first collecting all relevant vertices through a radius search on the centroid $c(f_c)$ with radius $2\Xrd$ using the vertex k-d-tree. 
The half-edge data structure allows us to access all the triangles involved with these vertices. 

\boldparagraph{Vertex Proximity Test.}
Vertex insertions very close to existing vertices are undesirable (Figure~4f) since it requires the insertion of very small faces at subsequent prediction steps.
We therefore consider triangle candidates close to an existing triangle as overlapping.
The threshold  $t_v = \frac{\defaultradius}{2}$ is applied along the axis defined by the face normal and on the distance to the triangle edges on the triangle plane. 
More formally, we define the test for whether point $\Xp$ overlaps with triangle $f$ as
\begin{equation}
\begin{aligned}
	\text{PIT}(\Xp, f)=\  d_{\text{vertical}} < t_{\text{near}}\quad \wedge 
	\min \{\Xsdf_{e_0}, \Xsdf_{e_1}, \Xsdf_{e_2} \} >- t_{\text{near}}
\end{aligned}
\end{equation}
where $d_{\text{vertical}}$ is the smallest distance between $\Xp$ and the triangle plane and $\Xsdf_e$ denotes the signed distance of $\Xp$ projected onto the triangle plane, to the lines defined by the triangle edges. Note that these values are positive if the point lies on the same site of the line as the triangle and negative otherwise. 

\boldparagraph{Segment Intersection Test.} 
Consider two edges $e_1$ and $e_2$, both belonging to different triangles. 
Let $\Xv_{e_1,\text{a}}$ and $\Xv_{e_1,\text{b}}$ denote the vertices of $e_1$ and $\Xv_{e_2,\text{a}}$ and $\Xv_{e_2,\text{b}}$ denote the vertices belonging to $e_2$.
Furthermore, let $\Xp_1$ and $\Xp_2$ be the points on edge $e_1$ and edge $e_2$, respectively, belonging to the shortest line segment which connects these two
lines.
The points can be expressed as 
\begin{equation}
\begin{aligned}
    \Xp_1&= \Xv_{e_1,\text{a}} + d_1 (\Xv_{e_1,\text{b}}-\Xv_{e_1,\text{a}})\\
    \Xp_2&= \Xv_{e_2,\text{a}} + d_2 (\Xv_{e_2,\text{b}}-\Xv_{e_2,\text{a}}),
\end{aligned}
\label{eq:segment_points}
\end{equation}
with $\Xp_1$ and $\Xp_2$ being on the triangle edge if $d_1 \in [0,1]$ and $d_2 \in [0,1]$, respectively.
Therefore, we consider two edges intersecting, if the two points $\Xp_1$ and $\Xp_2$ are located on the respective edges and the euclidean distance between the points is within a defined threshold, \ie $d_1, d_2 \in [0, 1]$ and $\norm{\Xp_2 - \Xp_1} \leq t_{\text{near}}$.

To compute the scalars $d_1$ and $d_2$, we make use of the constraint that the line segment is perpendicular to the edges. Hence, the dot product must be zero, \ie $(\Xv_{e_i,\text{b}}-\Xv_{e_i,\text{a}})^{\top} (\Xp_2-\Xp_1) =0$ for $i \in \{0,1\}$. By rewriting the vector $\Xp_2-\Xp_1$ with Eq.~\ref{eq:segment_points}, we are able to solve for the scalars $d_1$ and $d_2$.

\boldparagraph{Existing Vertex Selection.}
If one of the above tests indicates that the predicted vertex overlaps with existing triangles, the predicted vertex is replaced with an existing vertex $v_A$.
Only vertices returned from the radius search used to build the triangle set $F_{f_c}$ (Sec.~\ref{sec:overlap_details}) are considered as candidates.
Among the available candidate vertices, we choose the candidate with the smallest euclidean distance to the center of the boundary edge, provided the triangle defined by the vertex candidate does not overlap with the existing mesh. 
We therefore perform an additional overlap check for all candidate vertices.
Note that vertex proximity test must not be applied to faces containing the vertex candidate, since the two faces always intersect at the vertex candidate. 

\section{Curvature Head Details}
\label{sec:curvature_head_details}
\begin{figure}[t]
    \includegraphics[width=1.0\linewidth]{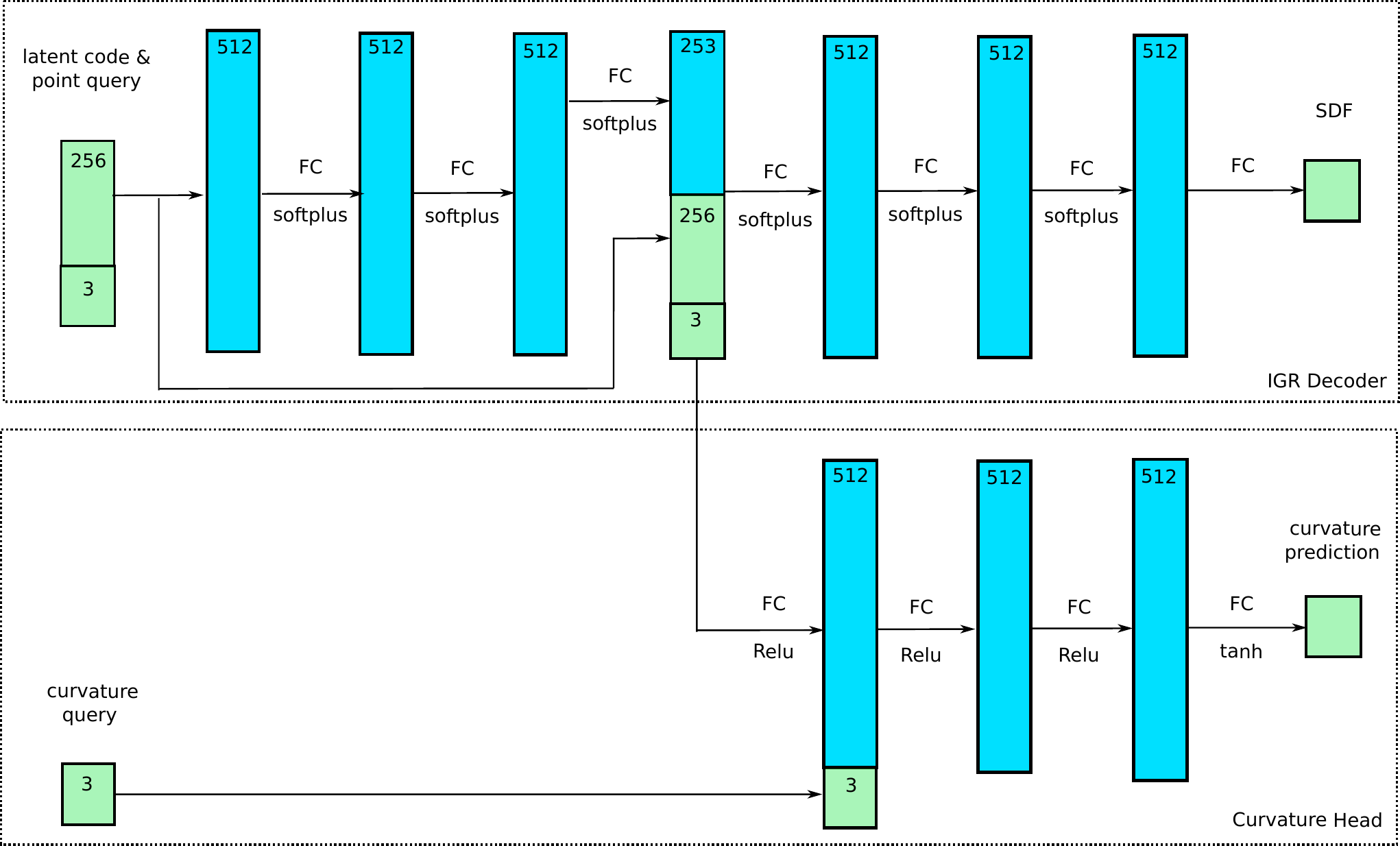}
    \caption{\textbf{Curvature head architecture.} The curvature head consists of 3 fully connected layers with ReLU activation functions. We attach the curvature head at the layer after the skip connection of the original IGR decoder. The directional curvature query is only provided to the curvature head
    }
    \label{fig:net_curvature_head}
\end{figure}
As shown in Figure~\ref{fig:net_curvature_head}, we attach the curvature head at the layer after the skip connection. 
The head consists of three fully connected layers of width 512 with ReLU activations, 
except for the output layer where we use tanh as the activation function. 
The directional curvature query $\Xqdir$ is exclusively provided to the curvature head and doesn't get used for standard SDF queries.
The decoupled curvature head allows us to use pretrained decoder weights from models trained without a curvature head. 
During training, the query direction $\Xqdir$ is randomly sampled on the tangential plane of each query point, defined by the surface normal. 

\section{Vertex Prediction Details}
\label{sec:vertex_predictor_details}
\boldparagraph{Architecture.}
\begin{figure}[t]
    \includegraphics[width=1.0\linewidth]{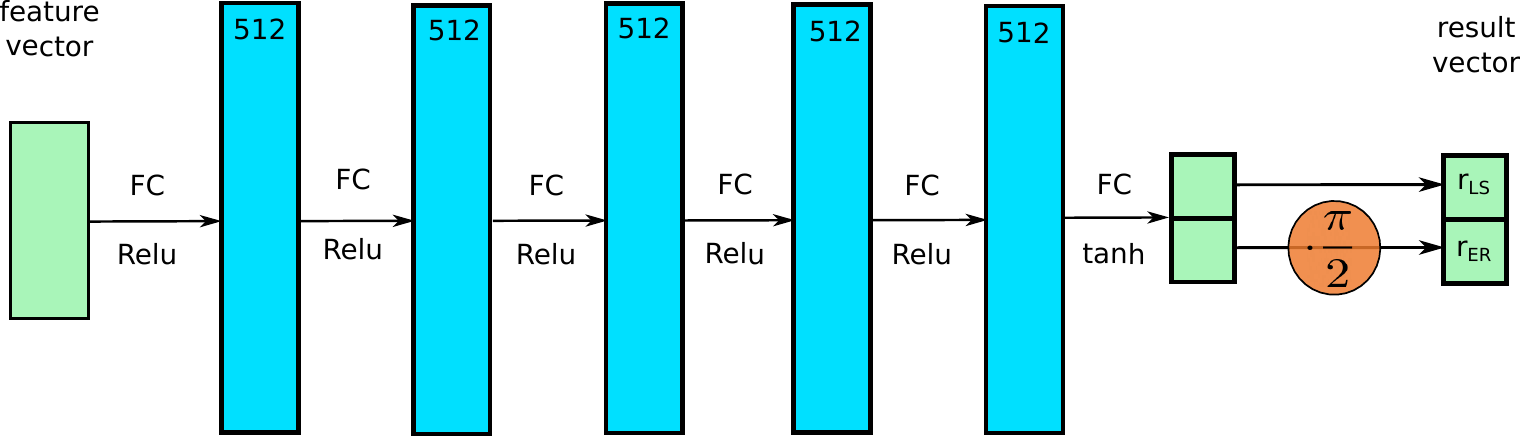}
    \caption{\textbf{Vertex prediction architecture.} The MLP for the vertex prediction module consists of 5 fully connected layers with ReLU activation functions. 
    The angular value in the result vector is scaled by $\pihalbe$
    }
    \label{fig:net_vp_mlp}
\end{figure}
The MLP in the vertex prediction module, shown in Figure \ref{fig:net_vp_mlp}, consists of five fully connected layers of width 512. We use ReLU as activation function and tanh for the output layer.
The angular value in the result vector is scaled by $\pihalbe$. 
\boldparagraph{Training.} 
We use the Adam optimizer~\cite{kingma_adam_2015} with an initial learning rate of $0.001$ 
without any learning rate scheduling. 
The MLP is trained iteratively with new batches (of 32 shapes) of triangle insertions.
Each shape is initialized with 1024 initialization triangles. 
A training step consists of a mini batch containing 512 boundary edges per shape. 
For every batch of 32 shapes, 50 training steps are computed.  
Currently, we train a model for every mesh resolution, defined by $\Xrd$.

\section{Data Preprocessing}
\label{sec:data_preprocessing}
\boldparagraph{Data Normalization.}
We normalize meshes and point clouds by centering them to the mean and scaling them to the unit cube, \ie to $[-1,1]$ for every dimension.
We perform this normalization for the training of the modified implicit representation and for all evaluations.

\boldparagraph{Artifact Removal.}
\begin{figure}[t]
    \includegraphics[width=1.0\linewidth]{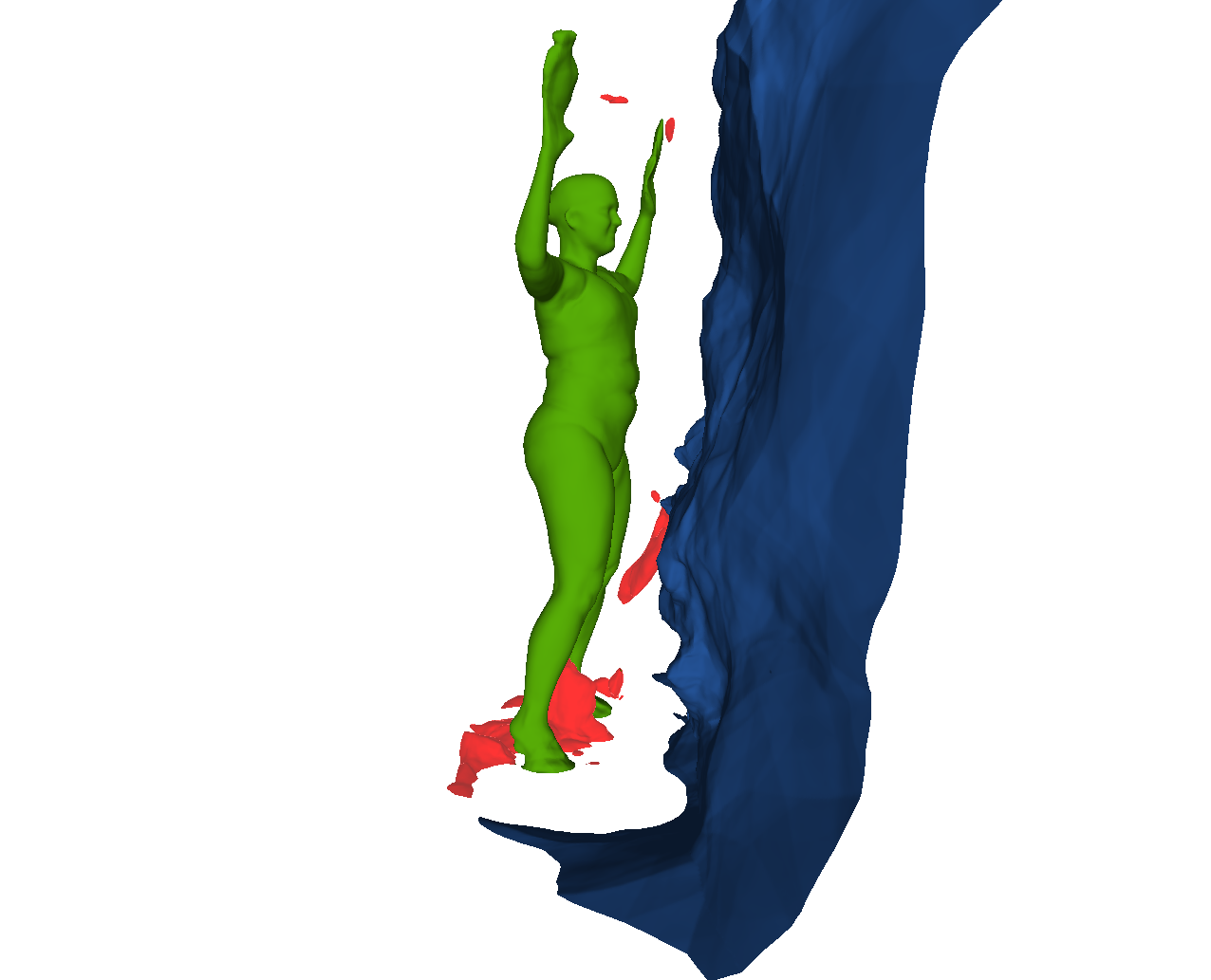}
    \caption{\textbf{Artifacts of implicit representations.} The original scan contains the human only (green). IGR introduces artifacts such as volumetric objects (red) around the body and a curtain in front of the person (blue). For evaluations, we remove such artifacts}
    \label{fig:igr_artefacts}
\end{figure}
The implicit neural representation (IGR) used by marching cubes and our method introduces several shape artifacts.
An example of such artifacts is visualized in Figure \ref{fig:igr_artefacts}. 
Furthermore, as the implicit representation is an approximation of the ground-truth mesh, it might differ slightly in some surface areas, smoothing sharp features or providing explanations for parts of the shape where no input data was provided.
In order to compare our method with approaches which do not use an implicit representation, \eg PointTriNet~\cite{sharp_pointtrinet_2020}, we remove such artifacts automatically.
The removal process first discards all triangles in the lowest five percent of the bounding box, 
\ie we cut of the feet of humans in the DFaust dataset since the artifacts often are connected with the scans through the ground and the feet.
We then split the mesh into connected components and keep only the component closest to the center of the bounding box. The distance from a mesh component to the center of the bounding box is computed by measuring the euclidean distance between the center of the bounding box and the mean of all mesh-vertices.

\section{More Quantitative Results}
\label{sec:quantitative_results}
\boldparagraph{Impact of Inlier Threshold on F1-score.}
\begin{figure}[t]
	\centering
    \captionsetup[subfigure]{aboveskip=-4pt,belowskip=-10pt}
	\begin{subfigure}[t]{0.49\linewidth}
		\centering
		\includegraphics[width=\linewidth]{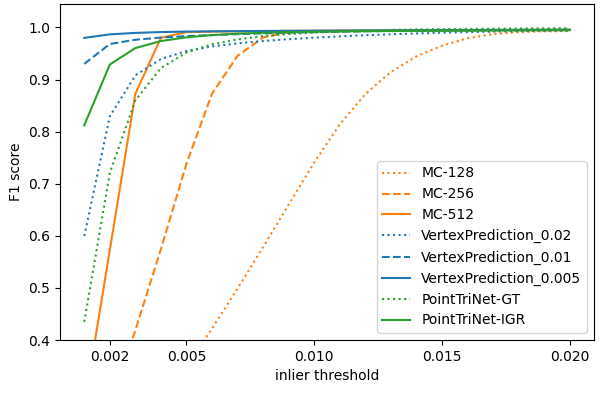}
		\caption{}
		\label{fig:f1plot_igr}
	\end{subfigure}
	\begin{subfigure}[t]{0.49\linewidth}
		\centering
		\includegraphics[width=\linewidth]{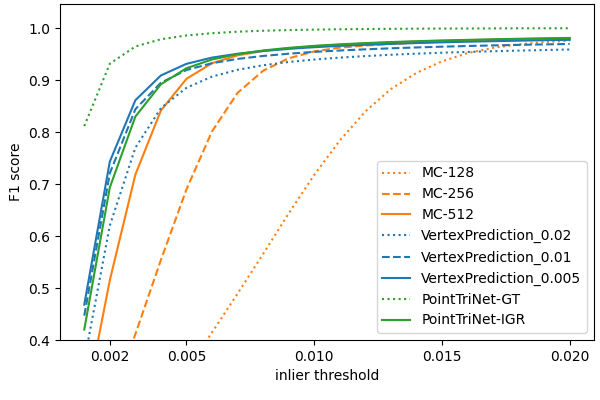}
		\caption{}
		\label{fig:f1plot_gt}
	\end{subfigure}
	\caption{\textbf{Impact of the inlier threshold on the F1-score.} (a) F1-score of the generated meshes measured on the implicit representation. (b) F1-score of the generated meshes measured on the ground-truth meshes}
	\label{fig:f1plots}
\end{figure}
To provide a better picture on the impact of the threshold on the F1-score, 
Figure~\ref{fig:f1plots} visualizes the change of the F1-score with growing $t_{\text{inlier}}$. 
Compared to marching cubes, our method achieves a better F1 score for similar resolutions with tight inlier thresholds. 
PointTriNet meshes are closer to the ground truth mesh, which we attribute to the artifacts and differences in the implicit representation.

\boldparagraph{Normal Consistency.}
\input{tab_normal_consistency}
To further evaluate the consistency of the orientation of the generated faces, we employ the normal consistency metric (NC).
The corresponding results are reported in Table~\ref{tab:results_dfaust_nc}.
The quality of the face orientation are within the same range for marching cubes and our method when evaluated on the implicit representation. 

\boldparagraph{Interaction with Implicit Representation.}
\input{tab_num_sdf_queries}
Marching cubes and \shortTitle{} rely on querying the underlying implicit representation. 
In Table~\ref{tab:eval_num_sdf_queries} we report the average number of queries performed in the evaluated meshing procedures on the D-Faust dataset. 
We can observe that the extensive querying performed by our algorithm is cheaper than the grid-based queries employed by marching cubes.

\section{More Qualitative Results}
\label{sec:more_qualitative_results}
\begin{figure}[t]
\centering
    \includegraphics[width=0.95\linewidth]{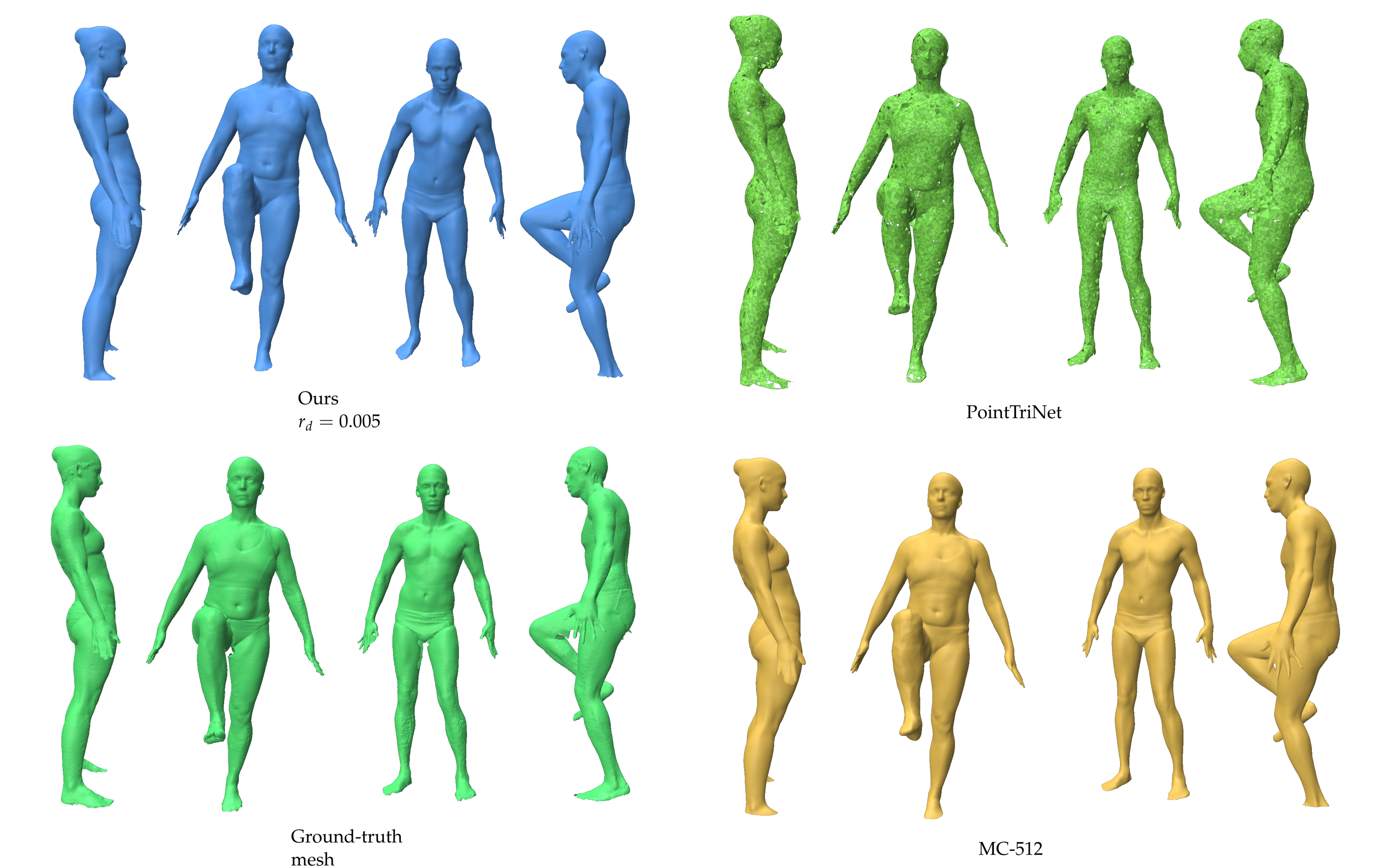}
    \caption{\textbf{More qualitative results on D-Faust.} We show 2 subjects in 2 different poses, respectively, and compare our version with $\Xrd=0.005$ with marching cubes~\cite{lewiner_efficient_2003} at $512$ resolution and PointTriNet~\cite{sharp_pointtrinet_2020}
    }
    \label{fig:more_qualitative}
\end{figure}
\begin{figure}[t]
\centering
    \includegraphics[width=0.95\linewidth]{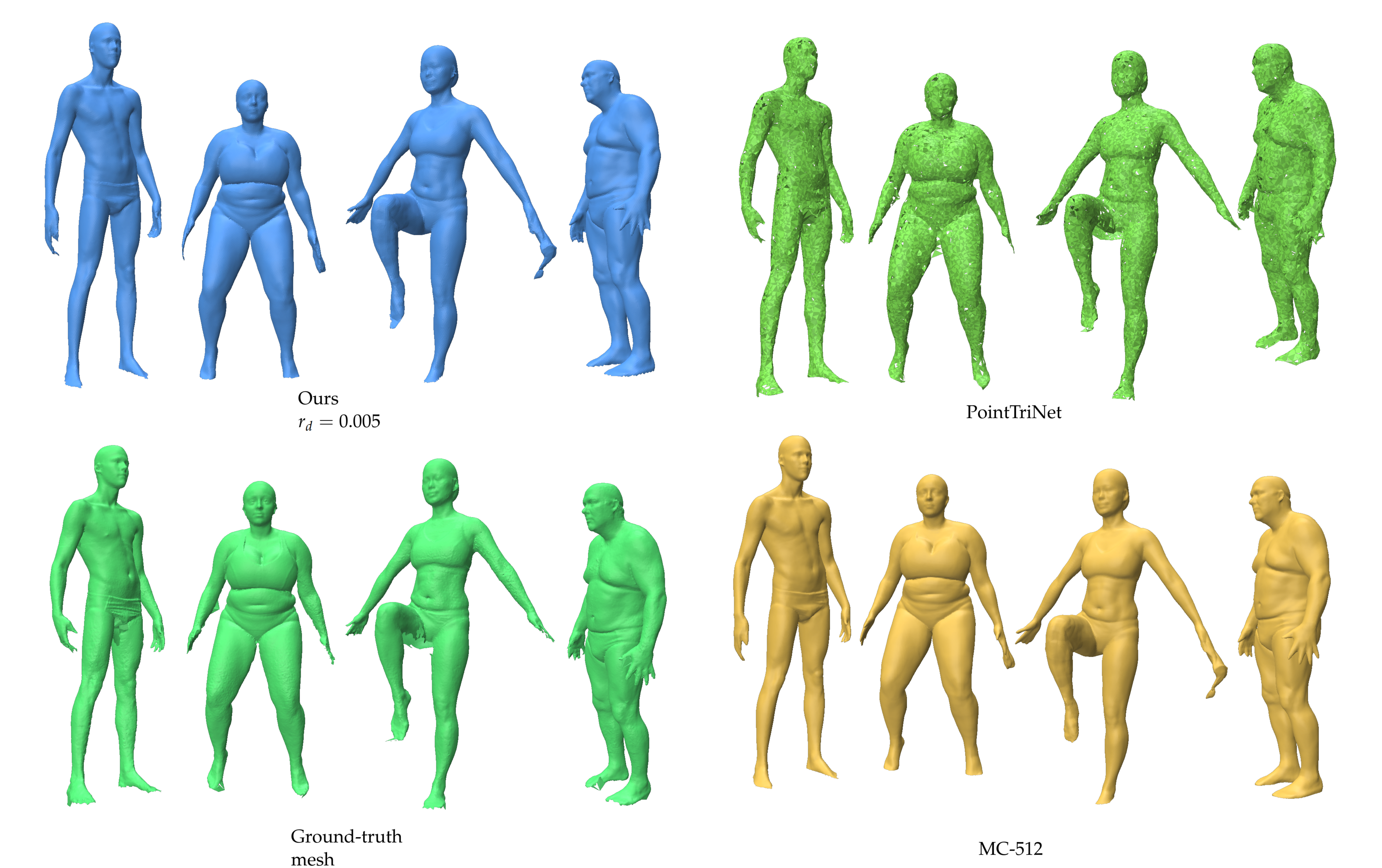}
    \caption{\textbf{More qualitative results on D-Faust.} We show 4 subjects in different poses, respectively, and compare our version with $\Xrd=0.005$ with marching cubes~\cite{lewiner_efficient_2003} at $512$ resolution and PointTriNet~\cite{sharp_pointtrinet_2020}
    }
    \label{fig:more_qualitative_2}
\end{figure}
In Figure~\ref{fig:more_qualitative} and Figure~\ref{fig:more_qualitative_2}, we provide additional qualitative results.
We can observe that our meshes are more complete than those from PointTriNet, while still providing the same level of detail as marching cubes.

\section{Ablation Study}
\label{sec:ablation_results}
\input{tab_ablation_distances}
To understand the impact of different elements of our meshing algorithm on the reconstruction quality, we perform an ablation study and report the results in Table~\ref{tab:ablation_chamfer_distances}.
For a given triangle size of $\Xrd=0.005$, we retrain the \textit{vertex predictor} with different components disabled, while using a similar experiment setup as before.
We run all ablations with a disabled surface projection post-processing step, as it otherwise skews the results, as differences between the ablations get tiny.
For the \textit{no length scaling} experiment, we use the default triangle size $\Xrd$ instead of the predicted length, while for the \textit{no edge rotation} experiment, we use a predicted angle of zero, corresponding to an extension of the mesh along the plane defined by the existing triangle adjacent to the boundary edge.
For the \textit{no prediction} experiment, we do not use either prediction.
The next 4 experiments test the effectiveness of each embedded feature. 
We alternatingly either omit the features at the default vertex (\textit{no $p$-queries}) or at the boundary edge end points (\textit{$p$-queries only}), use only 1 directional curvature feature queries instead of the original 3 (\textit{$1$ direction}) or use no curvature features at all (\textit{no curvature}).

\section{Watertightness}
\label{sec:watertightness}
\input{tab_hole_eval}
We report the boundary edge to total edge ratio in Table~\ref{tab:eval_hole_metrics} as a metric for watertightness. 
Since some shapes intersect the bounding box, marching cubes also introduces boundary edges at these locations.
\shortTitle{} exhibits fewer and smaller holes than PointTriNet while marching cubes does not generate any. 

\section{Limitations}
\label{sec:limitations}
\boldparagraph{Sharp features}
\begin{figure}[t]
	\centering
    \includegraphics[width=1.0\linewidth]{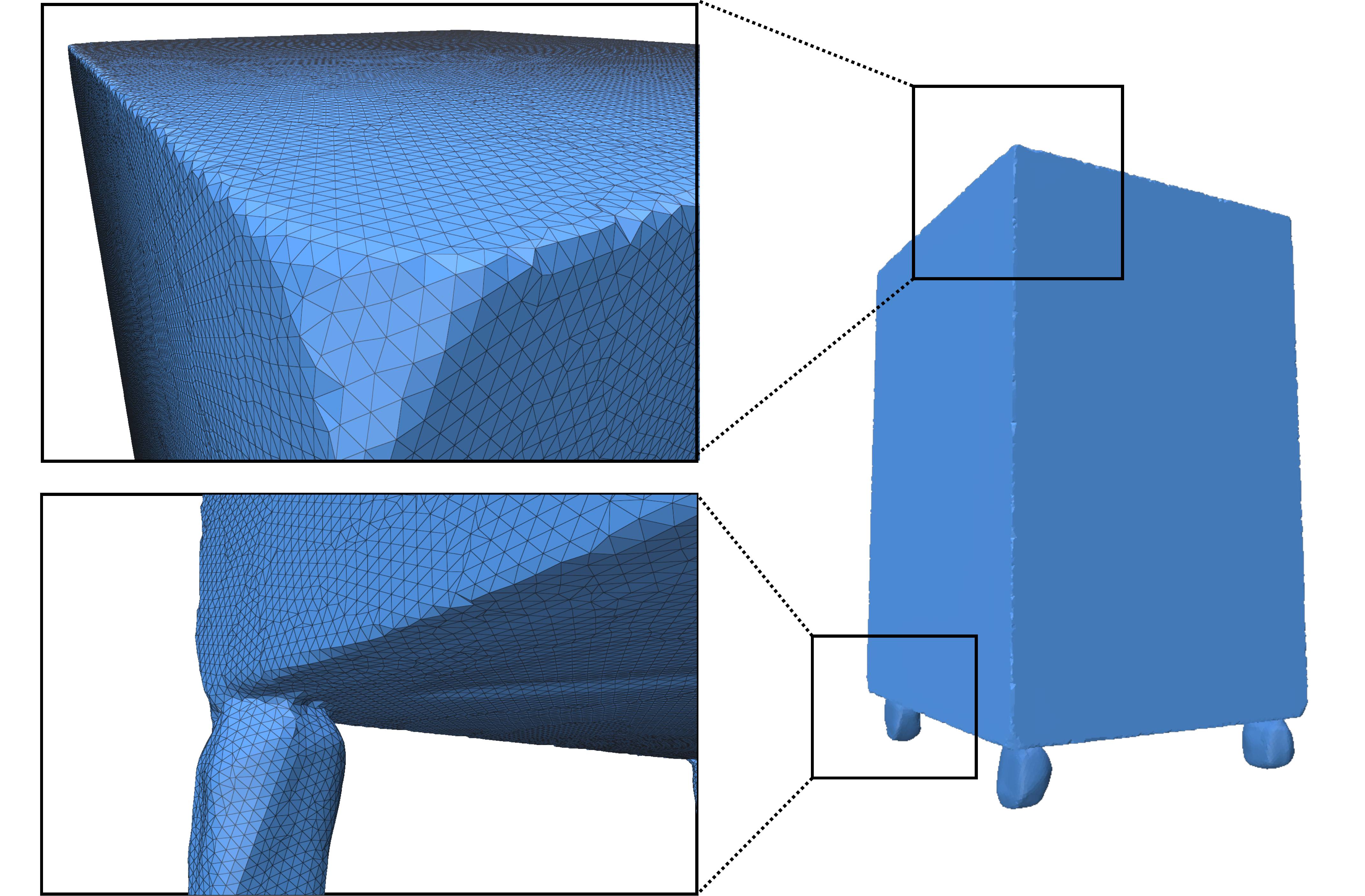}
    \caption{\textbf{Sharp features.} \shortTitle{} naturally extends to sharp edges and corners but smooths them due to the smoothing effects of the underlying implicit representation. Example belongs to the ShapeNet dataset
    }
    \label{fig:sharp_edges}
\end{figure}
The subset of the ShapeNet dataset used for evaluation contains features such as corners and sharp edges. \shortTitle{} naturally extends to sharp edges and corners, as visible in Figure~\ref{fig:sharp_edges}. 
However, due to the limited information provided to the \textit{vertex prediction module} and the smoothing effects of the underlying implicit representation, the meshing procedure reconstructs smoother corners and can lead to irregularities along sharp edges.

\begin{figure}[t]
	\centering
	\begin{subfigure}[t]{0.49\linewidth}
		\centering
		\includegraphics[width=\linewidth]{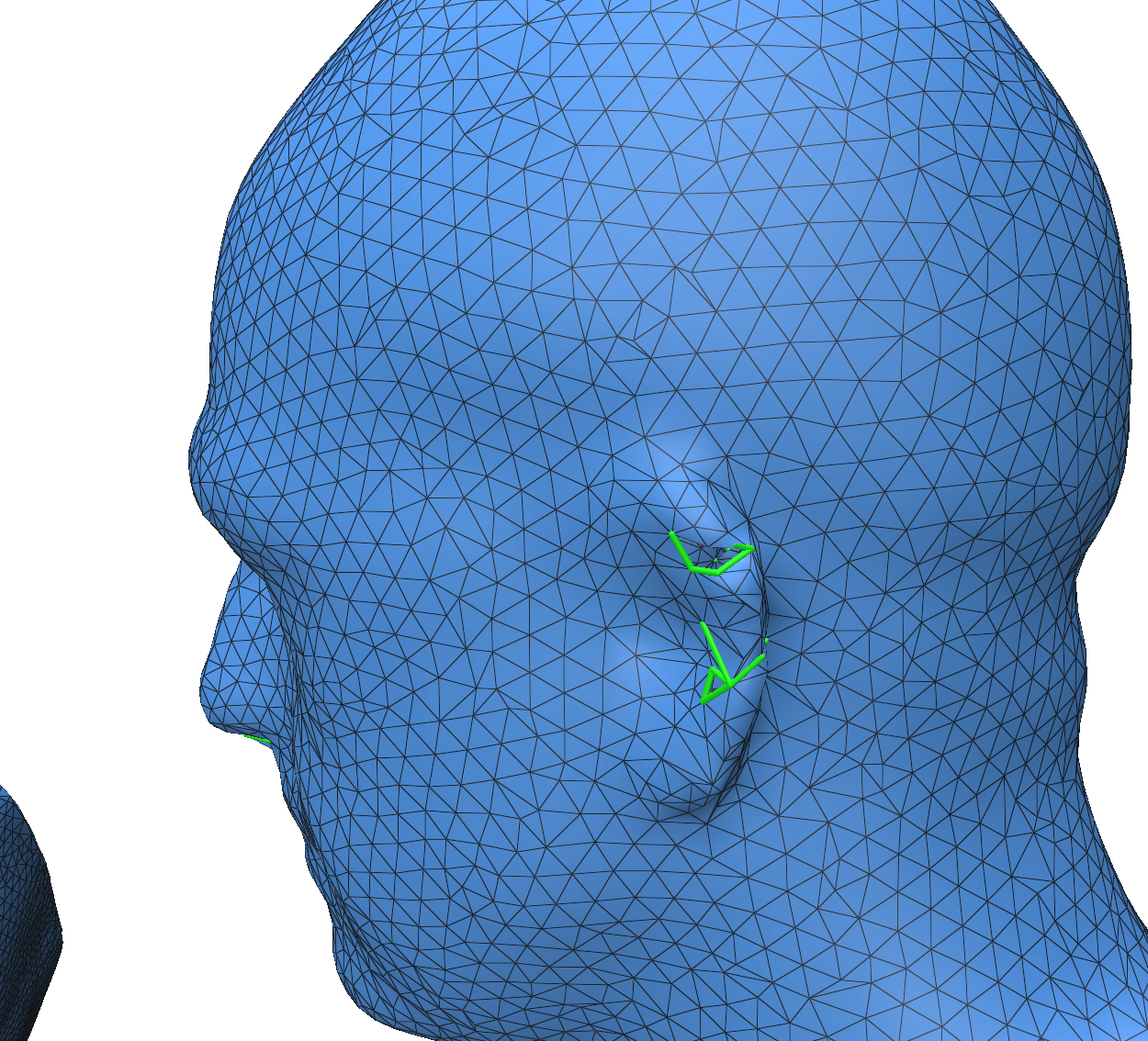}
		\caption{}
		\label{fig:failure_ear}
	\end{subfigure}
	\hfill
	\begin{subfigure}[t]{0.49\linewidth}
		\centering
		\includegraphics[width=\linewidth]{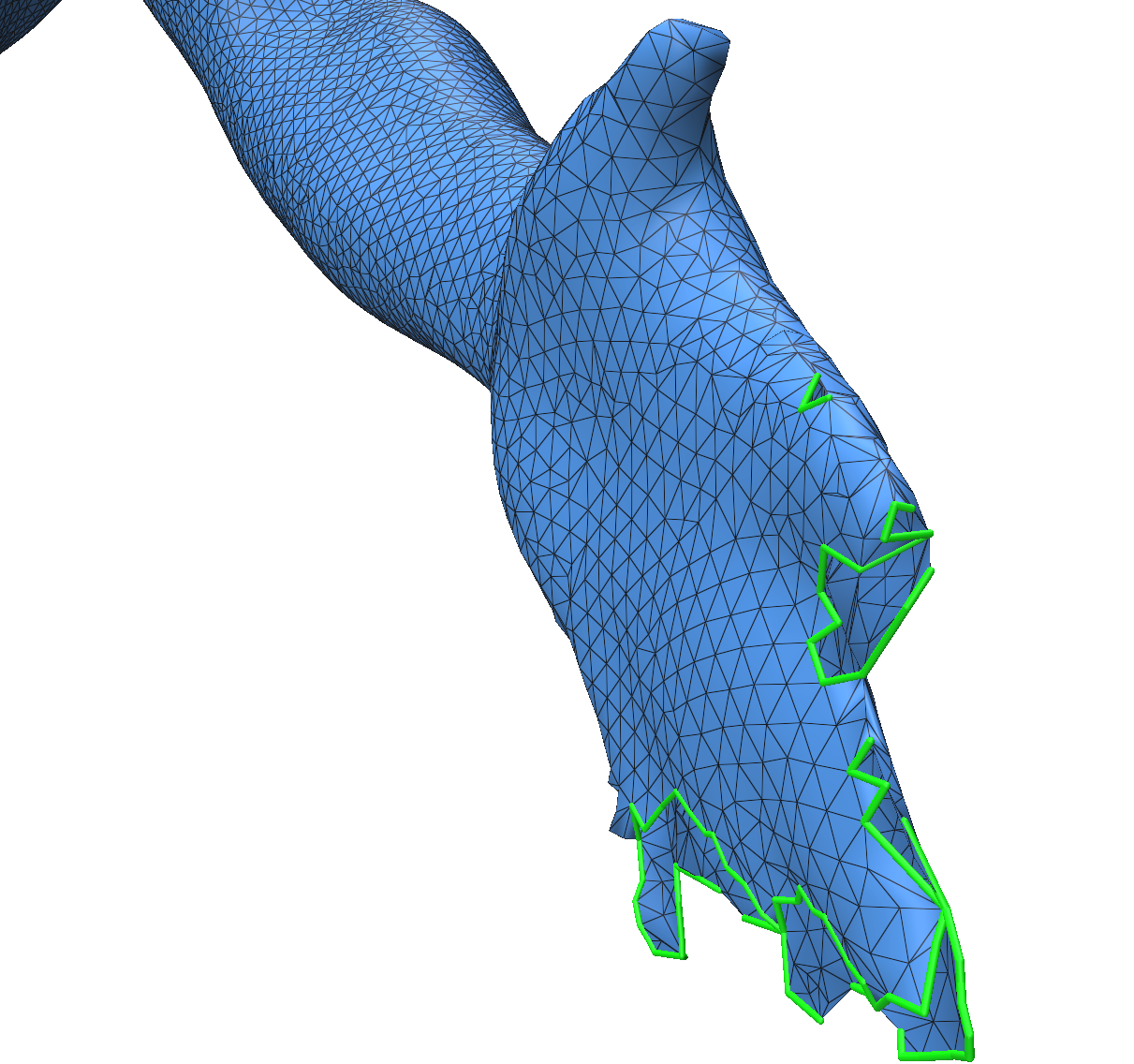}
		\caption{}
		\label{fig:failure_hand}
	\end{subfigure}
	\newline
	\begin{subfigure}[t]{0.49\linewidth}
		\centering
		\includegraphics[width=\linewidth]{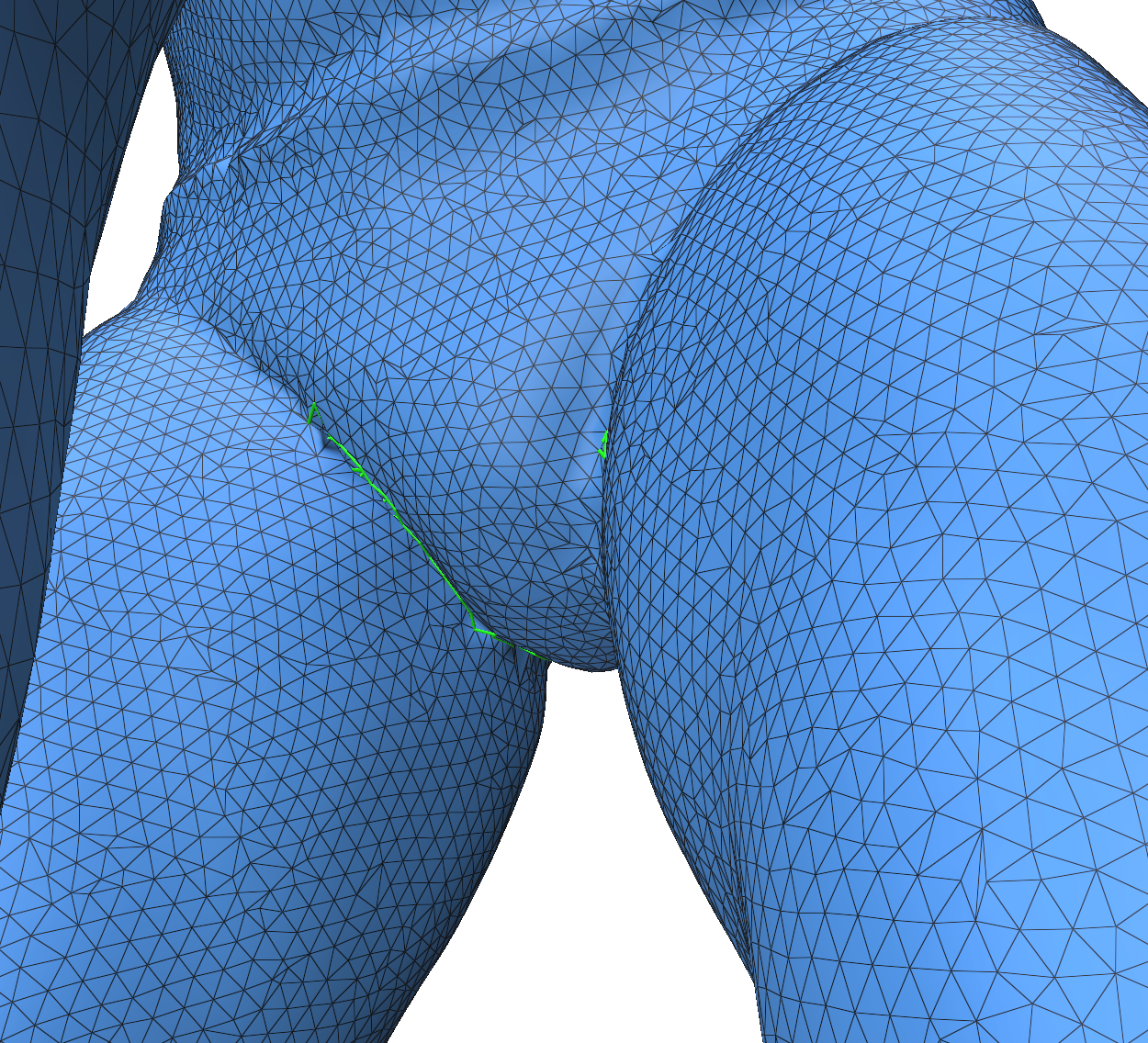}
		\caption{}
		\label{fig:failure_schritt}
	\end{subfigure}
	\hfill
	\begin{subfigure}[t]{0.49\linewidth}
		\centering
		\includegraphics[width=\linewidth]{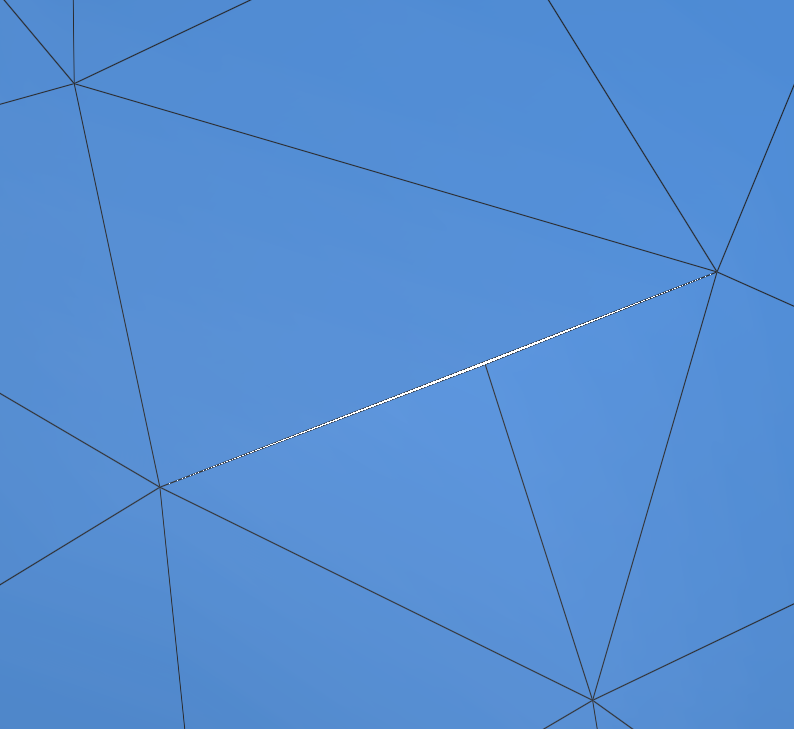}
		\caption{}
		\label{fig:failure_thin}
	\end{subfigure}
	   \caption{\textbf{Holes.} \shortTitle{} does not guarantee watertightness and 
	   therefore, holes may occur
	   on very complex surfaces such as ears (a), nearly parallel surface patches (b) or in places with large differences in orientation (c). Flat surfaces can lead to barely noticeable holes (d).
	   Green edges indicate boundary edges}
	   \label{fig:failure_cases}
\end{figure}

\boldparagraph{Holes.}
Holes occur mostly at surface locations with high curvature, illustrated in Figure~\ref{fig:failure_ear}, \ref{fig:failure_hand} and \ref{fig:failure_schritt}. 
In such cases, the edge rotation prediction does not bend the triangle sufficiently towards the 
real surface, leading to a scenario where no existing vertices are found, and the mesh therefore remains unclosed. 
Holes in flat surfaces can occur, but are rare and almost unnoticeable, as shown Figure \ref{fig:failure_thin}.

\boldparagraph{Supervision.}
The current method requires an explicit mesh for computation of the supervision signal of the curvature head. 
Future work would address this by incorporating self-supervision in order to preserve training on raw scans only.

\boldparagraph{Initialization Triangles.}
A single initialization triangle is sufficient for reconstructing a single connected component, but will fail for shapes containing multiple disconnected surface components.
The procedure must therefore sample enough points in order to catch all surface components with high probability. Such cases can easily be reduced by using the grid points of an initialisation grid instead of random points in space.

Furthermore, initialization triangles placed on a surface region with high curvature might
introduce inaccurate faces in the mesh. Future work could improve upon this by adapting the size of the initialization triangle to the curvature.

%% file: tab_normal_consistency.tex
\begin{table}[tb]
    \centering
    \footnotesize
    \caption{\textbf{Normal consistency.} The normal consistency of %
    generated meshes is measured against the implicit representation (IGR) and the ground-truth (GT)}
    \label{tab:results_dfaust_nc}
    \begin{tabular}{llcc}
    \multicolumn{2}{r}{} & NC on GT $\uparrow$ & NC on IGR $\uparrow$ \\
    \toprule\multirow{3}{*}{ Ours }
    & $\defaultradius=0.02$ & 0.466  & 0.942  \\
    & $\defaultradius=0.01$ & 0.479  & 0.967  \\
    & $\defaultradius=0.005$ & 0.485  & 0.980  \\
    
    \midrule
    \multirow{3}{*}{ MC \cite{lewiner_efficient_2003} }
    & $res=128$ & 0.475  & 0.953  \\
    & $res=256$ & 0.483  & 0.973  \\
    & $res=512$ & 0.486  & \B 0.981  \\
    
    \midrule
    \multicolumn{2}{l}{ PointTriNet GT~\cite{sharp_pointtrinet_2020}  } & \B 0.490  & 0.968  \\
    \multicolumn{2}{l}{ PointTriNet IGR~\cite{sharp_pointtrinet_2020}  } & 0.483  & 0.971  \\
    
    \end{tabular}
\end{table}

%% file: tab_num_sdf_queries.tex
\begin{table}[tb]
    \centering
        \caption{\textbf{Number of SDF queries.} We report the number of SDF queries used by marching cubes~\cite{lewiner_efficient_2003} and our method. \shortTitle{} uses significantly less SDF queries than marching cubes}
    \label{tab:eval_num_sdf_queries}
    \begin{tabular}{llr}
        {} & {} & \# SDF queries\\
        \midrule
        \multirow{3}{*}{Ours}
        & $\defaultradius=0.02$   & 264'534.06 \\
        & $\defaultradius=0.01$   & 1'059'887.04 \\
        & $\defaultradius=0.005$  & 3'488'435.31 \\
        \midrule
        \multirow{3}{*}{MC \cite{lewiner_efficient_2003}}
        & $res=128$               & 2'097'152.00 \\  
        & $res=256$               & 16'777'216.00 \\  
        & $res=512$               & 134'217'728.00 \\ 
    \end{tabular}

\end{table}

%% file: tab_ablation_distances.tex
\begin{table}[tb]
    \centering
    \caption{\textbf{Ablation.} In order to better distinguish the individual contributions of each component, we run all ablations without the surface projection post-processing step, except for the last line. The evaluations are performed in the same manner as in Table~1
    }
    \label{tab:ablation_chamfer_distances}
    \setlength{\tabcolsep}{3pt} %
    \renewcommand{\arraystretch}{1.05} %
    \resizebox{1.0\linewidth}{!}{
    \begin{tabular}{llrrrrrr}
    \multicolumn{2}{r}{from} & \multicolumn{2}{c}{Generated mesh} & GT & IGR & \multicolumn{2}{c}{Bidirectional} \\
    \cmidrule(lr){3-4}\cmidrule(lr){5-6}\cmidrule(lr){7-8}
    \multicolumn{2}{r}{to} & \gb{ 1e-4 } GT $\downarrow$ & \gb{ 1e-4 } IGR $\downarrow$ & \multicolumn{2}{c}{\gb{ 1e-4 } Generated Mesh $\downarrow$ } & \gb{ 1e-4 }GT $\downarrow$ & \gb{ 1e-4 } IGR $\downarrow$ \\
    \midrule
    \multicolumn{2}{l}{ No Edge Rotation  } &\B  24.943  & 16.154  & 143.354  &  138.992 & 84.149  & 77.573  \\
    \multicolumn{2}{l}{ No Length Scaling  } & 54.467  & 15.809  & 21.797  &  13.979 & 38.132  & 14.894  \\
    \multicolumn{2}{l}{ No Prediction  } & 25.222  & 16.121  & 150.392  &  146.241 & 87.807  & 81.181  \\
    \multicolumn{2}{l}{ No p-queries  } & 53.427  & 18.129  & 21.022  &  13.237 & 37.225  & 15.683  \\
    \multicolumn{2}{l}{ 1-direction  } & 55.127  & 17.551  & 21.587  &  14.170 & 38.357  & 15.861  \\
    \multicolumn{2}{l}{ p-queries only  } & 146.458  & 17.804  & 23.749  &  16.784 & 85.103  & 17.294  \\
    \multicolumn{2}{l}{ No Curvature  } & 148.378  & 16.097  & 22.172  &  14.335 & 85.275  & 15.216  \\
    \midrule
    \multicolumn{2}{l}{ Ours $\defaultradius=0.005$, w/o proj.  } & 55.720  & 15.318  & 21.971  &  13.847 & 38.846  & 14.583  \\
    \multicolumn{2}{l}{ Ours $\defaultradius=0.005$, w/ proj.  } & 44.763  &\B 1.093  & \B 19.967  & \B 5.627 & \B 32.365  & \B 3.360  \\
    \end{tabular}
    }
\end{table}

%% file: tab_hole_eval.tex
\begin{table}[tb]
    \centering
    \caption{\textbf{Quantitative metrics of observed holes.} 
    We report the ratio of boundary edges to total edges, the average number of holes per shape, the average radius of the minimum enclosing sphere of holes and the average number of edges involved in a hole
    }
    \label{tab:eval_hole_metrics}
    \scriptsize
    \setlength{\tabcolsep}{6pt} %
    \renewcommand{\arraystretch}{1.05} %
    \begin{tabular}{llrrrr}
    {} & {} & Ratio $\downarrow$ & \#Holes $\downarrow$ & Radius $\downarrow$& \#Edges $\downarrow$ \\
    \midrule\multirow{3}{*}{ Ours }
    & $\defaultradius=0.02$ & 0.022  & 12.438  & 0.005  &  15.447  \\
    & $\defaultradius=0.01$ & 0.004  & 14.344  & 0.002  &  16.749  \\
    & $\defaultradius=0.005$ & 0.003  & 19.125 & 0.001  &  19.574  \\
    
    \midrule
    \multirow{3}{*}{ MC \cite{lewiner_efficient_2003} }
    & $res=128$ & 0.002  & 1.781  & 0.010  &  35.965  \\
    & $res=256$ & 0.001  & 1.812  & 0.011  &  73.983  \\
    & $res=512$ & 0.001  & 1.781  & 0.011  &  151.754  \\
    
    \midrule
    \multicolumn{2}{l}{ PointTriNet GT \cite{sharp_pointtrinet_2020}  } &0.130  & 682.281  & 0.002  &  31.194  \\
    \multicolumn{2}{l}{ PSR \cite{kazhdan_poisson_2006}  } & \B 0.000  & \B 0.406  & \B 0.001  &  \B 9.077  \\
    
    \end{tabular}
\end{table}